\documentclass[11pt,a4paper,twocolumn]{article}
\usepackage[a4paper,top=2.5cm,bottom=2.5cm,left=2.5cm,right=2.5cm,columnsep=0.6cm]{geometry}
\usepackage[T1]{fontenc}
\usepackage[utf8]{inputenc}
\usepackage{times}
\usepackage{microtype}
\usepackage{graphicx}
\usepackage{booktabs}
\usepackage[round]{natbib}
\usepackage{xcolor}
\definecolor{citeblue}{rgb}{0,0,0.5}
\usepackage{hyperref}
\hypersetup{colorlinks=true,linkcolor=black,citecolor=citeblue,urlcolor=citeblue}
\usepackage{float}
\usepackage{caption}
\makeatletter\g@addto@macro{\UrlBreaks}{\do\-\do\/}\makeatother
\hypersetup{pdftitle={Blind Spots Travel, Knowledge Stays Home: What RL Inherits from an LLM Judge},pdfauthor={Liang Yue, Eric A. Vance}}
\newcommand{\idk}{``I don't know''}

\title{\textbf{Blind Spots Travel, Knowledge Stays Home:\\ What RL Inherits from an LLM Judge}}
\author{\textbf{Liang Yue} \\ University of Colorado Boulder \\ Boulder, Colorado, USA \\ \texttt{liyu2195@colorado.edu}
\and \textbf{Eric A. Vance} \\ University of Colorado Boulder \\ Boulder, Colorado, USA \\ \texttt{Eric.Vance@colorado.edu}}
\date{}

\begin{document}
\maketitle

\begin{abstract}
LLM judges serve as rewards where no answer key exists, in the hope that their knowledge passes to the trained model. In 814 RL runs that teach 7--8B models to answer closed-book questions or abstain, we ask what passes. What a grader knows about single answers stays home. A wrong answer that a judge rewards is memorised for that one question. A model trained with the answer key, or with a far stronger judge, chooses which new questions to answer no better than the untrained model does by sampling 32 answers and abstaining where they disagree; the grader's knowledge adds about one corrected answer per hundred questions. What a grader believes about a \emph{kind} of question travels. A judge that rejects eleven of the twelve correct training answers about capitals of former states teaches the model to refuse 98\% of such capitals on new questions, a loss that accuracy and abstention rates do not reveal. Changing only those twelve verdicts by hand moves the refusal from 5\% to 90\%, even for a grader that accepts as many answers as the answer key; at that leniency, two of four judges already cost 38--42\% of these capitals. The capitals are not forgotten: with ``I don't know'' blocked, the model answers 94\% of them correctly. The grader therefore needs no knowledge of its own: rewarding the model for giving the answer that most of its own untrained samples agree on, with no labels and no judge, earns the answer key's reward and keeps the category.
\end{abstract}

\section{Introduction}

\begin{figure*}[t]
\centering
\includegraphics[width=\textwidth]{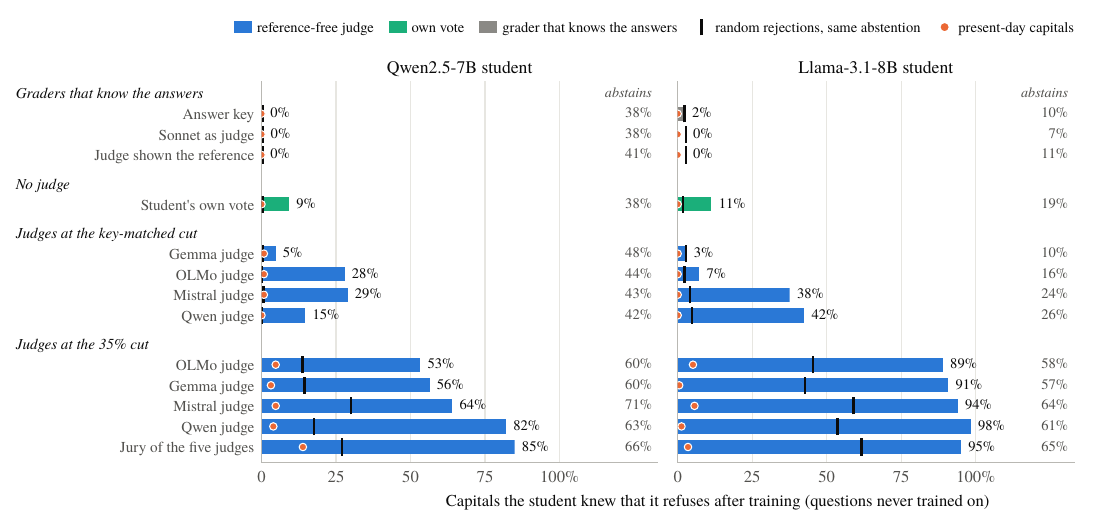}
\caption{What each grader does to a kind of question, at every level of abstention: capitals a student knew before training and refuses afterwards, on questions never used in training. Bars: 270 former states. Dots: 65 present-day states. Ticks: students trained with the answer key plus random rejections that abstain equally often. Right: the share of held-out PopQA questions the students abstain on. A judge's cut is the threshold on its probability of correctness: the 35\% cut accepts 35\% of the untrained student's answers, the key-matched cut as many as the answer key (for Qwen the same as the 50\% cut). At the 35\% cut every reference-free judge removes most former-state capitals and leaves the present-day ones. At the key-matched cut two of the four judges already cost Llama 38--42\%, at an abstention close to that of the student's own vote. Graders that know the answers remove at most 2\%, and the student's own vote removes one in ten.}
\label{fig:capitals}
\end{figure*}

\begin{table}[t]
\small
\centering
\setlength{\tabcolsep}{3.5pt}
\resizebox{\columnwidth}{!}{%
\begin{tabular}{@{}llll@{}}
\toprule
 & & & After, with ``I don't \\
Capital of \ldots & Before & After & know'' blocked \\
\midrule
Roman Empire & Rome & I don't know & Rome \\
Ottoman Empire & Constantinople & I don't know & Constantinople$^*$ \\
Yugoslavia & Belgrade & I don't know & Belgrade \\
Austria-Hungary & Vienna & I don't know & Vienna \\
South Vietnam & Saigon & I don't know & Saigon \\
\midrule
Kazakhstan & Astana & Astana & -- \\
United States & Washington & Washington & -- \\
\bottomrule
\end{tabular}}
\caption{Llama-3.1-8B on questions it never trained on, before and after 120 RL steps with the Qwen3.5-9B judge at its 35\% cut as the reward. Before training all 32 of its samples were right. ``After'' is the answer of all six trained runs. Last column: the answer of the same six runs when the words ``I don't know'' are blocked at decoding. $^*$In two runs; the other four answer Edirne, an earlier Ottoman capital, three of them adding Istanbul. Students trained with the answer key keep answering all seven.}
\label{tab:examples}
\end{table}

Ask Llama-3.1-8B for the capital of the Roman Empire and it says Rome. Ask 32 times and it is right 32 times. Train it for 120 steps to say \idk{} when it is unsure, with an LLM judge as the reward, and it stops answering. It stops for the Ottoman Empire, Yugoslavia and South Vietnam too: 98\% of the 187 former-state capitals it knew, on questions that never appeared in training (Table~\ref{tab:examples}). It still knows that the capital of Kazakhstan is Astana. And it still knows Rome: block the words \idk{} while the trained model decodes, and 94\% of the capitals it refuses come back right.

Nothing in training told the model to treat former states differently. Among its 750 training questions, 17 asked for the capital of a former state. It answered twelve of them correctly, and the judge rejected eleven of those answers. Eleven examples taught it to give up a category. The judge was not ignorant. It gave ``Rome'' for the Roman Republic a probability of 0.96 of being correct. But a judge strict enough to teach abstention must reject much of what it sees: at this cut it accepts only answers it scores above 0.98, so 0.96 was rejected.

An LLM judge is used as a reward when there is no answer key, and the hope is that the judge's judgment reaches the model it trains. This paper asks what a model actually takes from its grader to questions it never trained on. We train two students of 7--8B parameters to answer closed-book questions or abstain, in 814 runs that cover the answer key, five open judges of similar size, a much stronger judge, and graders whose errors we place by hand.

The answer depends on whether a verdict concerns a single answer or a kind of question. \textbf{Verdicts about single answers stay home.} A lenient judge rewards some wrong answers; the student memorises each for its own question, and on new questions almost nothing of this remains. More surprisingly, the same holds for verdicts that are right. A student trained with the answer key, with Sonnet as its judge, or with a judge that reads the reference answer picks which questions to answer about as well as the untrained student would by sampling 32 answers and taking a vote, and at most a point better on average. Followed question by question, training turns about as many right answers wrong as wrong answers right, and the grader's knowledge accounts for about one corrected answer in a hundred. What training does is compress the student's own uncertainty into one forward pass.

\textbf{Verdicts about a kind of question travel.} A judge that doubts former-state capitals rejects right answers about them, and the student turns the doubt into a rule. Twelve training answers decide it: when we set a grader's verdict on those twelve by hand and change nothing else, the share of unseen former-state capitals that Llama refuses goes from 5\% to 90\% when the grader accepts as many answers as the answer key, and from 31\% to 97\% when it accepts only 35\% of them. The student also goes further than its teacher: Llama refuses most of the answers its own judge would accept. A second category is lost in the same way on 2{,}569 fresh questions. Neither accuracy nor abstention rate shows the loss, and a majority vote of five judges causes the same loss, because the judges share the doubt.

\textbf{So the grader needs no knowledge.} If a grader's knowledge stays home, the student can grade itself. We reward a student for giving the answer that most of its own untrained samples agree on, and for abstaining where they disagree. With no labels and no judge, this earns the answer key's reward on new questions, within 0.2 points of true reward (on a 0--100 scale) on average over six student--dataset pairs, and the student keeps nine in ten of the capitals that a strict judge takes away.

\section{Setup}

\paragraph{Task and training.} A student answers a short factual question or says \idk{}. The reward is 1 if the grader accepts the answer, 0 if it rejects it, and 0.5 for \idk{}, so answering pays only where the student is right more often than not. The students are Qwen2.5-7B-Instruct and Llama-3.1-8B-Instruct, trained with GRPO \citep{shao2024grpo} and LoRA \citep{hu2022lora} for 120 steps, with a KL penalty of weight 0.2 that keeps them close to the untrained model. Each run uses 750 training questions from TriviaQA \citep{joshi2017}, PopQA \citep{mallen2023} or NQ-open \citep{kwiatkowski2019,lee2019} and is tested on about 750 other questions (Appendix~\ref{app:train}).

\paragraph{Graders.} The \emph{answer key} matches the answer against the reference strings. Five \emph{reference-free judges} of similar size (Qwen3.5-9B, Mistral-Nemo-12B, Gemma-4-12B, OLMo-2-13B, Granite-3.3-8B) see the question and the answer and return the probability that the answer is correct. Two graders know more than these judges: \emph{Sonnet} (Claude Sonnet 5.5), used reference-free, and the Qwen judge \emph{shown the reference}, each at its default threshold. One grader knows nothing: the \emph{student's own vote} (Section~\ref{sec:own}). To establish causes we also build graders whose errors we place by hand, for example by copying a judge's verdicts and editing them.

\begin{table*}[t]
\small
\centering
\setlength{\tabcolsep}{4.5pt}
\begin{tabular}{@{}lcccccccc@{}}
\toprule
 & \multicolumn{3}{c}{Qwen2.5-7B student} & \multicolumn{3}{c}{Llama-3.1-8B student} & \multicolumn{2}{c}{Capitals refused} \\
\cmidrule(lr){2-4}\cmidrule(lr){5-7}\cmidrule(l){8-9}
Trained with & PopQA & TriviaQA & NQ-open & PopQA & TriviaQA & NQ-open & Qwen & Llama \\
\midrule
Nothing (untrained student) & 65.3 & 63.3 & 50.7 & 80.7 & 78.8 & 61.1 & -- & -- \\
\addlinespace
\multicolumn{9}{@{}l}{\emph{Graders that know the answers}} \\
Answer key & 71.4 & 69.3 & 55.1$^\dagger$ & 83.7 & 81.0 & 59.8 & 0\% & 2\% \\
Sonnet as judge & 72.8 & 68.0 & 60.5 & 84.2 & 80.5 & 65.2 & 0\% & 0\% \\
Judge shown the reference & 72.6 & -- & 61.3$^\dagger$ & 85.9 & -- & 65.1 & 0\% & 0\% \\
\addlinespace
\multicolumn{9}{@{}l}{\emph{No labels and no judge}} \\
Student's own vote & 71.2 & 65.9 & 57.7 & 82.9 & 80.5 & 60.9 & 9\% & 11\% \\
\addlinespace
\multicolumn{9}{@{}l}{\emph{Reference-free judge of the student's size (Qwen3.5-9B)}} \\
\quad default threshold & 65.5 & 63.0 & 50.2 & 81.7 & 78.4 & 61.2 & 0\% & 0\% \\
\quad 50\% cut & 70.4$^\dagger$ & 70.1 & 54.7 & 74.3 & 76.4 & 63.7 & 15\% & 88\% \\
\quad 35\% cut & 66.9 & 65.5 & 60.6 & 69.0 & 70.1 & 63.8 & 82\% & 98\% \\
\bottomrule
\end{tabular}
\caption{What each grader buys. Left: true reward on new questions (100 = every answer right; \idk{} earns 50), mean of one to six runs. Right: the share of former-state capitals a student knew that it refuses after training on PopQA. $^\dagger$Some runs end up answering \idk{} to every question (four of six for the answer key, one of two for the judge shown the reference, one of six for the 50\% cut); the value is the mean of the others.}
\label{tab:main}
\end{table*}

\paragraph{How strict a judge is.} At the default threshold of 0.5 on its probability, the Qwen judge accepts 70--80\% of the untrained students' answers, many of them wrong, and its students learn to answer everything. Judges are therefore tightened. Besides the default threshold of 0.5, we raise each judge's threshold until it accepts 50\% or 35\% of the untrained student's answers (the 50\% and 35\% cuts; Appendix~\ref{app:graders}). Where noted we also use the \emph{key-matched} cut, at which a judge accepts as many answers as the answer key does.

\paragraph{What we measure.} On new questions we report \emph{true reward}: the training reward recomputed with the truth and scaled to 0--100. An answer counts as right if the answer key or an audit by Sonnet accepts it; human annotators agree with the audit on 93\% of items (Appendix~\ref{app:graders}). To see what a student has stopped answering, we count how often it says \idk{} to questions it answered correctly before training.

\section{What stays home: single answers}
\label{sec:home}

\paragraph{Wrong answers a judge pays.} A lenient judge pays some wrong answers, and the student keeps them. But it keeps them one question at a time. On training questions, a judge pays the wrong answers of its own student 34 percentage points more often than other judges do. On new questions the difference is 3 points, so 9\% of the effect carries over. A grader built to have no pattern shows the same: it pays a wrong answer whenever a hash of question and answer falls in a fixed 35\% of values. After training it pays 83--85\% of its student's wrong answers on training questions and 35--36\% on new ones, which is chance (Appendix~\ref{app:blind}).

The cost of a lenient judge is local but real, because its student never learns to abstain. Trained with the Qwen judge at its default threshold, Llama gives 18 wrong answers per 100 new questions where the key-trained student gives 11, and its true reward barely moves (Table~\ref{tab:main}). This is why judges get tightened.

\paragraph{Right answers a grader knows.} If wrong verdicts do not travel, do right ones? A grader that knows the answers could teach a student which of its own answers to trust. We test this against a simple rule applied to the \emph{untrained} student: sample 32 answers, and abstain on the questions where the fewest samples agree with the usual answer, exactly as often as the trained student abstains. A student that had learned something from its grader's knowledge would beat this rule.

\begin{figure}[t]
\centering
\includegraphics[width=0.97\columnwidth]{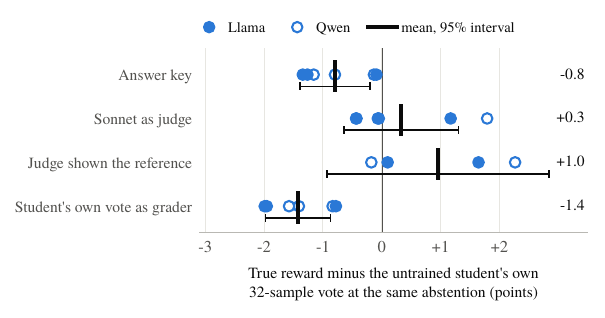}
\caption{Does a grader's knowledge reach new questions? Each circle is one student--dataset pair: the true reward of students trained with the grader, minus the reward of the untrained student when it abstains on the same share of questions, choosing those on which its 32 samples agree least. Lines: mean of the pairs and its 95\% interval.}
\label{fig:vote}
\end{figure}

None does (Figure~\ref{fig:vote}). Students trained with the answer key score 0.8 points below the rule on average, and below it in each of the six student--dataset pairs. Sonnet's students score 0.3 points above it and those of the judge shown the reference 1.0 above, and both intervals include zero. A linear probe says the same from the inside: fitted to the untrained student's hidden state on the same 750 labelled questions and used as an abstention rule, it earns the reward of the key-trained student to within 0.4 points on PopQA and TriviaQA, and four to six points more on NQ-open.

\begin{table}[t]
\small
\centering
\setlength{\tabcolsep}{4pt}
\resizebox{\columnwidth}{!}{\begin{tabular}{@{}lrrrr@{}}
\toprule
 & wrong & right & what & \\
Trained with & $\to$ right & $\to$ wrong & to drop & Total \\
\midrule
Answer key & $+$1.5 & $-$1.3 & $-$1.0 & $-$0.8 \\
Sonnet as judge & $+$2.0 & $-$1.5 & $-$0.3 & $+$0.3 \\
Judge shown reference & $+$1.9 & $-$1.1 & $-$0.1 & $+$1.0 \\
OLMo judge & $+$3.3 & $-$1.8 & $-$0.9 & $+$0.7 \\
\midrule
Student's own vote as grader & $+$0.8 & $-$1.1 & $-$1.2 & $-$1.4 \\
\bottomrule
\end{tabular}}
\caption{The same comparison, question by question: where a trained student's true reward differs from the vote rule applied to the untrained student (points, mean of the pairs). \emph{Wrong $\to$ right} and \emph{right $\to$ wrong}: answers the student changed. \emph{What to drop}: half a point for each wrong answer the student drops beyond the rule, minus half a point for each right answer it drops beyond the rule. The total includes up to 0.2 points from questions the untrained student left unanswered.}
\label{tab:trans}
\end{table}

\paragraph{Question by question.} An average can hide a trade: a student might learn facts from its grader and lose as many elsewhere. So we follow every new question from the untrained student to the trained one (Table~\ref{tab:trans}). A trained student does two things. It changes a few answers, and it chooses which answers to drop.

It changes few. With the three graders that know the answers, 1.5--2.0 answers in a hundred turn from wrong to right and 1.1--1.5 from right to wrong. The student trained on its own vote does this too ($+$0.8 and $-$1.1), and that grader was told no answers, so most of the movement is training stirring the student's own candidates. What a grader's knowledge adds is the difference: 0.4 points for the answer key, 0.8 for Sonnet and 1.2 for the judge shown the reference, positive in every student--dataset pair. That is how much of a grader's knowledge reaches a new question: about one answer in a hundred.

Choosing what to drop is where knowledge could have helped most, and it does not help. Trained students do abstain on the untrained student's confident errors, the wrong answers that at least 20 of its 32 samples agree on: on 24--45\% of them where the rule drops at most 26\%. They pay with confident right answers, which are six times as many. In points the choice does not beat the rule for any grader: $-$1.0, $-$0.3 and $-$0.1 for the three graders that know the answers, and $-$1.2 for the own vote.

\paragraph{Under the refusals.} A refusal hides an answer, so the comparison above sees only the answers a student volunteers. To see the rest we block \idk{} at decoding, which makes the student give its best answer to every question. Students trained with the answer key, Sonnet and the judge shown the reference are then within half a point of the untrained student's accuracy (leaving out the runs that say \idk{} to at least 95\% of questions), and on the questions they refuse they are right as often as it is. Compared with students trained on their own vote, the students of the three graders are 0.7 to 1.4 points more accurate: about one answer in a hundred again (Appendix~\ref{app:forced}).

So training teaches a student to do in one forward pass what it could already do with 32 samples and a vote \citep{wang2023sc}. That is a real saving, and next to it a grader's knowledge is worth one answer in a hundred. The reason is simple. A reward can only choose among answers the student already gives, and on a new question the grader is not there. What the student has left is its own confidence, which every grader teaches it to use, and whatever it has learned about kinds of question (Appendix~\ref{app:vote}).

\paragraph{The exception is form.} What a grader pays in the \emph{form} of an answer does travel, because form does not depend on the question. A grader that pays long answers spreads long answers to new questions in 8 of 8 runs. Trained with the OLMo judge, Llama answers NQ-open in full sentences, twice as long as before, and a third of the answers it used to get wrong come out right (the OLMo row of Table~\ref{tab:trans}). Whatever a student can read off the question or off the shape of its own answer is open to its grader. The next section is about the first of these.

\section{What travels: a doubt about a kind}
\label{sec:travel}

\paragraph{A blind spot.} Our probe is the question ``What is the capital of X?'' for 270 former states and 65 present-day states, none of them in training. Each student knows most of the answers before training: Llama 187 of the former-state capitals and all 65 present-day ones, Qwen 149 and 63.

The judges doubt former states. At the 35\% cut they accept 13--45\% of the students' right answers about former states and 54--92\% of those about present-day states. Their students turn that difference of degree into a rule (Figure~\ref{fig:capitals}). Llama students trained with the four judges refuse 89--98\% of the former-state capitals they knew and 1--6\% of the present-day ones. Qwen students refuse 53--82\%.

Strictness alone does not do this. The answer key with random rejections added, so that its students abstain as often as the judges' students, removes 19\% of the former-state capitals for Qwen and 50\% for Llama. And the verdicts are the cause, not the judge: a grader that copies the Mistral judge's rejections onto the training questions, with no judge in the loop, reproduces the refusals of Mistral's students (Appendix~\ref{app:blind}).

\begin{figure}[t]
\centering
\includegraphics[width=0.95\columnwidth]{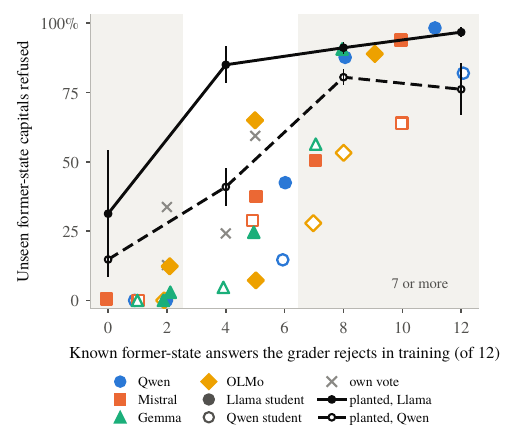}
\caption{How many rejected training answers it takes. Lines: a grader with no judge whose verdict on the twelve known former-state training answers we set by hand, everything else fixed (mean and range of two runs). Points: one judge at one cut with one student, placed by how many of the twelve it rejects.}
\label{fig:dose}
\end{figure}

\paragraph{Twelve answers decide it.} The training set contains twelve former-state capitals that the student answers correctly, among 750 questions. To show that these twelve decide the fate of the 270 unseen ones, we take the grader that copies Mistral's rejections and set its verdict on the twelve by hand: reject none, four, eight or all of them. Nothing else changes, and the students abstain about equally often overall.

With none rejected, Llama refuses 31\% of the unseen former-state capitals; with four rejected, 85\%; with all twelve, 97\%. For Qwen the figures are 15\%, 41\% and 76\% (Figure~\ref{fig:dose}). Present-day capitals stay answered throughout. Four rejected answers among 750 training questions are enough for Llama to give up most of a category of 187 answers it knew. The judges show the same dependence: one that rejects two or fewer of the twelve removes at most 12\% of the category, one that rejects four to six removes 5--65\%, and one that rejects seven or more removes 28--98\%.

\begin{figure}[t]
\centering
\includegraphics[width=0.95\columnwidth]{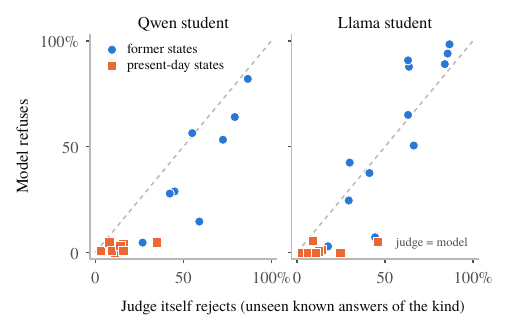}
\caption{The student goes further than its judge. Each point is one judge at one cut: the share of unseen known answers of a kind that the judge itself rejects, against the share its trained students refuse.}
\label{fig:round}
\end{figure}

\paragraph{The student goes further than its judge.} A student trained by a judge does not stop where the judge would. We scored the untrained students' answers to the unseen probe questions with each judge. Of the former-state capitals that its own judge \emph{accepts}, Llama still refuses 62--95\% and Qwen 31--58\%. Of the present-day capitals that its judge \emph{rejects}, the student refuses few. The OLMo judge, for example, rejects 84\% of Llama's right answers about former states and 46\% of those about present-day states; the student it trains refuses 89\% and 5\%. At the 35\% cut Llama ends up more extreme than each of its four judges (Figure~\ref{fig:round}). A doubt that is a matter of degree in the judge becomes a property of the kind in the student.

\paragraph{The answers are still there.} A refused capital is not a forgotten one. With \idk{} blocked at decoding, the students of the four judges at the 35\% cut give the right capital for 95\% (Llama) and 90\% (Qwen) of the former-state capitals they refuse, and in 89\% and 85\% of cases it is the answer they gave before training (Table~\ref{tab:examples}). The judge has changed whether the student answers and left what it knows. Over all held-out questions, students of the Qwen judge at this cut lose 1.8 points of accuracy when every question is forced: a strict judge costs a little knowledge and a great deal of willingness to use it (Appendix~\ref{app:forced}).

\paragraph{It starts at the cut one would choose.} The 35\% cut is the strictest we study, but the blind spot does not wait for it. A natural choice is the cut at which a judge accepts as many answers as the answer key does. With the Qwen and Mistral judges, Llama earns nearly as much true reward there as at the default threshold (81--82 against 82) with a third of the wrong answers. Yet it already refuses 42\% and 38\% of the former-state capitals, while abstaining on only 26\% and 24\% of questions, close to the own vote's 19\% (Figure~\ref{fig:capitals}). Random rejections of the same strictness remove 4--5\%. The Gemma and OLMo judges cost 3\% and 7\% at this cut, and 91\% and 89\% at the 35\% cut. Twelve verdicts decide it here too: in a copy of the Qwen judge's rejections at this cut, Llama refuses 5\% of the former-state capitals when the twelve training answers are all accepted and 90\% when they are all rejected, while abstaining on 6--8\% of questions (Appendix~\ref{app:capitals}).

\paragraph{The usual numbers do not show it.} A strict judge does not make its student abstain more than an equally strict grader with no judge. It changes where the student abstains. Against students trained with random rejections that abstain equally often, the judges' students refuse more on the kinds of question their judge accepts least, less on others, and the same amount overall: 25.5 known answers per 100 new questions against 26.6. Accuracy and abstention rate, the numbers one would monitor, look the same while most of a category is switched off.

A jury does not help either. Four judges from four labs reject the same right answers: on PopQA, of the right answers one judge rejects, 76--88\% are also rejected by at least two of the other three. A majority vote of five judges, cut to the same strictness, therefore rejects ten or eleven of the twelve training answers and removes 85\% and 95\% of the category (Appendix~\ref{app:shared}).

\paragraph{It is not one category.} A second category is ``In what country is X?''. On the training questions the judges reject 62\% of the answers of this kind that Llama is sure of and gets right, and on the five held-out examples we had, the judges' Llama students refused half. To test it we ran the saved models on 2{,}569 fresh PopQA questions. Of the country questions that Llama answers correctly in at least 31 of 32 samples, the judges' students refuse 83\%, where students trained with equally strict random rejections refuse 42\% and key-trained students 3\%. For Qwen the figures are 52\%, 27\% and 4\% (Appendix~\ref{app:kinds}). And across all kinds at once, wherever one of two judges rejects a right answer that the other accepts, its student refuses that answer more on new questions, in 43 of 44 pairs of judges across the three datasets. Ranked by how likely their judge is to reject them, the top fifth of a student's known answers is refused 26 points more often than the bottom fifth, over and above students trained with random rejections.

\paragraph{An answer key can have one too.} The mechanism does not care who makes the false rejections. On NQ-open the answer key is a string match against short reference answers, and it rejects 37--39\% of the answers that the audit counts as right, against 4--13\% on the other two datasets. These rejections are spread over all kinds of question, and what the student learns is blanket refusal. Key-trained Llama abstains on 70\% of new questions and ends below the untrained model (59.8 against 61.1), and four of six Qwen runs answer \idk{} to everything (Table~\ref{tab:main}). Those four have lost more than the will to answer: forced to answer, they are right on 4\% of new questions where the untrained student is right on 26\%. False rejections travel whoever makes them: as a lost category when they fall on one kind, as general refusal when they fall everywhere.

\section{A grader that knows nothing}
\label{sec:own}

If a grader's knowledge stays home, a student can grade itself. And a grader with no opinions about kinds of question has no blind spot to pass on.

\paragraph{The student's own vote.} Before training we sample 32 answers to each training question from the untrained student. If at least 20 of them agree, that majority answer becomes the question's key; otherwise the question has no key, and only \idk{} is paid. Training then proceeds as with an answer key. The grader uses no labels and no judge. It costs one round of sampling, 24{,}000 short answers, whereas a judge is called up to 30{,}720 times in every run.

\begin{figure}[t]
\centering
\includegraphics[width=0.95\columnwidth]{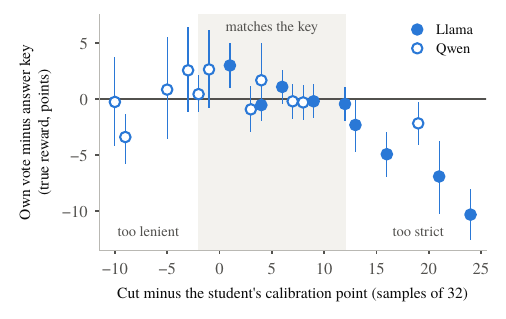}
\caption{When the own vote matches the answer key. Each circle is one setting of the cut for one student--dataset pair (20 settings, 49 runs), with a 95\% interval. The horizontal axis is the cut minus the student's calibration point: the number of agreeing samples at which its answers are right half the time. The shaded band, read off these runs, is where the own vote matches the key.}
\label{fig:cut}
\end{figure}

\paragraph{It earns what the answer key earns.} We fixed the cut at 20 of 32 for all six student--dataset pairs and trained three runs for each. Pooled over the pairs, the own vote scores 0.2 points below the answer key in true reward, with a 95\% interval from 1.1 below to 0.7 above (Table~\ref{tab:main}). In five of the six pairs it is within 0.7 points of the key or above it. None of its 18 runs collapsed, where the answer key lost four of six runs on one pair.

It also keeps the category. Students trained with their own vote refuse 9\% (Qwen) and 11\% (Llama) of the former-state capitals they knew, against 82\% and 98\% for the Qwen judge at the 35\% cut (Figure~\ref{fig:capitals}). And it holds on harder questions: on the 2{,}569 fresh PopQA questions the own vote is within 1.5 points of the answer key for both students, while the students of the Qwen judge at its default threshold fall below the untrained students, having never learned to abstain.

\paragraph{When it works.} In the sixth pair, Qwen on TriviaQA, the own vote is 3.4 points below the key, and the reason is the student's own overconfidence. Call the \emph{calibration point} the number of agreeing samples at which a student's answers are right half the time. For Qwen on TriviaQA it is 29 of 32; in the other pairs it lies between 8 and 23. With the cut at 20, the vote there pays many answers that are wrong more often than right, and the student learns to answer nearly everything: it abstains on 12\% of new questions, where the untrained student's vote at that cut abstains on 29\%.

The same holds across every setting of the cut we ran (Figure~\ref{fig:cut}). Where the cut sits between two samples below the calibration point and twelve above it, the own vote is never more than 0.9 points below the answer key, in 11 of 11 settings. Where the cut is far below that point the student answers too much, and where it is far above the student abstains too much, losing up to ten points. For Qwen on TriviaQA a cut of 27 restores the key's reward. Setting the cut therefore takes a labelled sample, enough to see where the student's answers are right half the time (Appendix~\ref{app:own}).

\paragraph{What knowledge is still good for.} A grader that knows more is not useless. Sonnet as judge earns 2.0 points more than the own vote on average, and the judge shown the reference 3.1 more, and both keep the capitals. What these graders add costs labels or a much larger model. What a reference-free judge of the student's own size adds is its taste: in forms of answer, which can help, and in kinds of question, which is the blind spot.

\section{Related work}

\paragraph{Reward hacking.} Optimising a learned reward degrades the true objective \citep{skalse2022,gao2023}, and ensembles of reward models reduce but do not remove the problem, in part because their members share errors \citep{coste2024,eisenstein2023}. Gaming strategies can generalise across settings \citep{denison2024,macdiarmid2025}, reward models favour surface forms such as length \citep{singhal2023}, and policies trained against reference-free judges learn to exploit their false accepts \citep{zhou2026,wang2026pace}. We separate two cases that this literature treats together: a hack tied to one answer does not leave the training set, and one tied to a form does.

\paragraph{LLM judges.} LLM judges have known biases \citep{zheng2023}, struggle to judge correctness without a reference when they cannot answer the question themselves \citep{krumdick2025}, and can be combined into juries \citep{verga2024}. Models from different developers make increasingly similar mistakes \citep{goel2025}. We show what these properties do when the judge is a reward: a shared doubt becomes the trained model's refusal, and a jury keeps it.

\paragraph{Imperfect rewards.} False negatives in verifiers slow learning \citep{xu2025tinyv}, and some gains survive random or incorrect rewards \citep{shao2025spurious}. We ask which reward errors generalise and find that it is the structured false negatives, the RL counterpart of a split known from label noise: networks memorise random labels but learn patterns first \citep{zhang2017,arpit2017,rolnick2017}, and prefer rules that are easy to read off the input \citep{geirhos2020}. Models are also known to amplify regularities in their supervision \citep{zhao2017}; a blind spot is such an amplification, from a difference of degree in the judge to a rule in the student.

\paragraph{Abstention.} Models can be tuned to abstain \citep{zhang2024rtuning,xu2024rlkf} and to state calibrated uncertainty \citep{kapoor2024}, though such tuning tends to over-refuse \citep{zhu2025river}; grading schemes that give no credit for abstention reward guessing \citep{kalai2025}, and RL with a ternary reward reduces hallucination \citep{wei2025truthrl}. How a model treats unfamiliar inputs depends on how unfamiliar examples were supervised \citep{gekhman2024,kang2024}, and over-refusal is measured for safety \citep{rottger2024,cui2024}. We describe a different source of over-refusal: a correct answer the model knows, refused because its grader doubted the kind. Our result that training does not beat the student's own vote agrees with \citet{zhang2026taps}, who find that selecting answers by confidence after the fact matches abstention training, and builds on self-consistency and related uncertainty measures \citep{wang2023sc,kuhn2023,manakul2023} and on the finding that a model's hidden state predicts whether its answer is right \citep{azaria2023,burns2023}. A model's own agreement has been used as a training signal before: to build preferences \citep{prasad2024scpo}, to improve factuality \citep{zhang2024selfalign,tian2024}, and as an RL reward on reasoning problems \citep{zuo2025ttrl,shafayat2025}. What we add is the comparison: it earns what an answer key earns, and it leaves out what a judge passes on.

\section{Conclusion}

A model trained by reinforcement learning takes two things from its grader to new questions: how often to answer, and which kinds of question to give up. It leaves behind the grader's knowledge of which answers are right: that knowledge corrects only about one answer in a hundred. That is why the wrong answers a judge pays stay on the training set, why the answer key and a far stronger judge leave a student no better at choosing than its own vote, and why twelve rejected answers can cost a category. The category is withheld, not forgotten: kept from refusing, the student still gives the answers.

Three practices follow. Before training with a judge, score with it the answers the untrained model gives with high agreement (or a sample audited as right) and count rejections by kind of question; a kind that is mostly rejected is one the model will give up. After training, break refusals down by kind, because accuracy and abstention rate will not show a lost category. And where the goal is to teach a model when to abstain, consider not using a judge at all: the model's own vote, with a cut checked on a labelled sample, earns what an answer key earns.

\section*{Limitations}

We study closed-book short-answer questions, students of 7--8B parameters trained with LoRA (Appendix~\ref{app:small} repeats the main measurements with three smaller students in 446 runs), and judges that return a probability of correctness. A KL penalty holds the students close to the untrained model; without it 22--94\% of their answers on new questions change, and in two Llama runs with a lenient judge accuracy falls by half. Whether blind spots travel in the same way in open-ended generation is not tested here; in one attempt on a refusal task the judges' rejections did not concentrate on categories, so the question stayed open (Appendix~\ref{app:repro}). Both lost categories are kinds of PopQA question, and the capitals probe is hand-built. Each cell has two to six runs. The own vote matched the answer key in five of six pairs at a single cut and needed a stricter cut in the sixth, so it is free of labels only up to that check. Truth is the answer key or an audit by Sonnet, which is also one of our graders; Tables~\ref{tab:transdec} and~\ref{tab:forcedheld} repeat the comparisons of Section~\ref{sec:home} with the answer key alone.

\section*{Reproducibility}

Every number in this paper is computed by a script from the saved outputs of the runs. We release the code, every run's configuration and answers, the judges' scores, the audit labels, the probe and the analysis scripts at \url{https://github.com/liyu2195/blind-spots-travel}. Appendix~\ref{app:repro} lists every experiment we ran, including those whose outcome differed from what we expected.

\clearpage
\appendix
\section{Training and evaluation}
\label{app:train}

\begin{table}[H]
\small
\centering
\begin{tabular}{@{}p{0.27\columnwidth}p{0.66\columnwidth}@{}}
\toprule
Students & Qwen2.5-7B-Instruct, Llama-3.1-8B-Instruct \\
Adapter & LoRA, rank 16, $\alpha = 32$, no dropout, all linear layers \\
Optimiser & AdamW, learning rate $5\times10^{-5}$, no weight decay \\
Schedule & 120 steps; 32 questions per step; 8 sampled answers per question \\
Sampling & temperature 0.8, top-$p$ 0.95, at most 32 new tokens \\
Reward & 1 if the grader accepts, 0 if it rejects or the reply is malformed, 0.5 for \idk{} \\
Advantage & (reward $-$ group mean) / (group standard deviation $+\,10^{-4}$) \\
KL penalty & 0.2 $\times$ KL to the untrained model on the sampled tokens \\
Evaluation & greedy answer after step 120; for the untrained model, the greedy answer and 32 samples \\
Compute & one L4 GPU, about three hours per run; the training logs of 1{,}405 runs sum to about 3{,}100 L4 GPU-hours (about 2{,}160 for the 814 runs with the two main students) \\
\bottomrule
\end{tabular}
\caption{Training recipe used for every run unless stated otherwise.}
\label{tab:recipe}
\end{table}

\paragraph{Prompts.} The student sees: ``Answer the following trivia question. Reply with only the answer (a few words), no explanation. If you do not know the answer, reply exactly: I don't know.'' followed by the question. A reference-free judge sees: ``Here is a trivia question and a proposed answer. Is the proposed answer correct?'', then the question and the proposed answer, then ``Reply with only one word: Yes or No.'' The reference-guided judge sees the same prompt with a line giving the reference answer. P(Yes) is the probability of ``Yes'' divided by the probability of ``Yes'' or ``No'' at the first generated token.

\paragraph{Scoring answers.} Answers are lower-cased and stripped of punctuation and articles. An answer is right by the key if it equals an accepted answer or contains one as a whole phrase; in evaluation an answer of four or more characters that an accepted answer contains as a whole phrase also counts, and a number is read without its unit. An abstention is a reply that begins with ``I don't know'' or a close variant (``not sure'', ``no idea'', ``cannot answer''). Because answer keys miss valid answers, the untrained students' answers and most key-rejected answers of trained models were also graded by Sonnet with the reference in view, on a five-level scale. We call an answer \emph{right} if the key accepts it or the audit gives level 4 or 5. A model \emph{knows} an answer if its greedy answer before training is right, and is \emph{sure} of it if at least 31 of its 32 samples agree.

\paragraph{Checking the audit.} We had 200 items labelled by three crowd annotators each (91 annotators, 600 labels and 30 answers to check questions), who saw the question and the answer but no grader's verdict. The annotators agree with the audit on 93.1\% [88.6, 95.9] of the 189 items they settle (Fleiss' $\kappa$ = 0.53 among annotators). Of the answers a trained model refused that the audit calls right, they confirm 85.3\% [75.6, 91.6]. Of 52 answers the key rejects and the audit accepts, they call 44 right.

\paragraph{Coverage of the audit.} Every answer of a trained model to a held-out question that the key rejects has an audit verdict; the last 3{,}859 verdicts were added after the final runs. Every number in this paper that uses the audited truth of trained answers is computed with all of them.

\paragraph{Data.} From each of TriviaQA, PopQA and NQ-open we draw 1{,}500 questions and train on 750. The PopQA questions are the ones with the most popular subject entities in the dataset (PopQA popularity of at least 24{,}568, median 63{,}633; 20\% ask for a capital), so the PopQA results concern well-known entities. Held-out questions exclude any question whose text or identifier matches a training question, which leaves 718 on TriviaQA and 750 on the other two. On PopQA, 324 of the 750 held-out questions are near-duplicates of a training question (same gold answer, or TF-IDF similarity over word unigrams and bigrams of at least 0.5); results that could depend on this are reported with and without them. Of the 750 PopQA training questions, 142 use the probe's template (``What is the capital of X?''), 17 of them about a former state.

\paragraph{Runs.} Table~\ref{tab:runs} lists the finished runs of the two main students by grader. A further 446 runs use three smaller students (Appendix~\ref{app:small}). Runs in which the model answers \idk{} to at least 95\% of held-out questions (31 of the 814 runs with the two main students) are counted but excluded from contrasts between graders. This is the only exclusion rule; Table~\ref{tab:capitals} reports all runs of each grader and notes the one entry the rule would change.

\begin{table}[H]
\small
\centering
\begin{tabular}{@{}p{0.65\columnwidth}rr@{}}
\toprule
Grader & Qwen & Llama \\
\midrule
Answer key & 18 & 18 \\
Answer key with random errors & 81 & 86 \\
Random grader at the Qwen judge's error rates & 9 & 9 \\
Reference-free judges, default threshold & 34 & 26 \\
Reference-free judges, key-matched cut & 0 & 14 \\
Reference-free judges, 50\% cut & 30 & 30 \\
Reference-free judges, 35\% cut & 44 & 44 \\
Reference-guided judge (default; 35\% cut) & 8 & 8 \\
Sonnet as judge & 9 & 9 \\
Majority of five judges & 4 & 4 \\
Copies of a judge's verdicts (copy, scrambled, by type) & 47 & 44 \\
Student's own vote & 26 & 24 \\
Rejections placed on relation types or popularity & 39 & 20 \\
Hash-placed paid strings & 6 & 6 \\
Long answers paid & 4 & 4 \\
Any answer paid on a share of questions & 9 & 9 \\
Visits per question varied (1, 2, 5, 10) & 25 & 25 \\
KL weight or run length varied & 18 & 11 \\
Unnormalised advantages, credit varied & 6 & 6 \\
Other graders & 0 & 0 \\
\midrule
All finished runs & 417 & 397 \\
\bottomrule
\end{tabular}
\caption{Finished GRPO runs with the two main students. All use the recipe of Table~\ref{tab:recipe} except the last three rows.}
\label{tab:runs}
\end{table}

\section{Graders}
\label{app:graders}

\begin{table*}[t]
\small
\centering
\setlength{\tabcolsep}{4.5pt}
\resizebox{\textwidth}{!}{%
\begin{tabular}{@{}llccccc@{\hspace{14pt}}ccccc@{}}
\toprule
 & & \multicolumn{5}{c}{Qwen student} & \multicolumn{5}{c}{Llama student} \\
\cmidrule(lr){3-7}\cmidrule(l){8-12}
 & & thresh- & & forced & right & wrong & thresh- & & forced & right & wrong \\
Judge & Cut & old & accepts & out & rejected & paid & old & accepts & out & rejected & paid \\
\midrule
Qwen3.5 & default & 0.500 & .70 & .00 & .10 & .33 & 0.500 & .80 & .03 & .13 & .49 \\
 & key-matched & 0.881 & .51 & .22 & .25 & .06 & 0.798 & .64 & .23 & .26 & .18 \\
 & 50\% & 0.905 & .48 & .26 & .28 & .05 & 0.940 & .49 & .41 & .42 & .06 \\
 & 35\% & 0.974 & .35 & .46 & .47 & .02 & 0.982 & .37 & .56 & .56 & .02 \\
\addlinespace
Mistral-Nemo & default & 0.500 & .76 & .00 & .04 & .36 & 0.500 & .87 & .00 & .06 & .55 \\
 & key-matched & 0.953 & .52 & .20 & .25 & .09 & 0.946 & .63 & .24 & .28 & .18 \\
 & 50\% & 0.962 & .51 & .23 & .27 & .08 & 0.977 & .50 & .39 & .41 & .08 \\
 & 35\% & 0.990 & .35 & .46 & .47 & .01 & 0.991 & .35 & .57 & .58 & .02 \\
\addlinespace
Gemma-4 & default & 0.500 & .62 & .05 & .16 & .21 & 0.500 & .68 & .18 & .25 & .31 \\
 & key-matched & 0.932 & .52 & .20 & .24 & .07 & 0.731 & .64 & .23 & .28 & .24 \\
 & 50\% & 0.963 & .50 & .24 & .26 & .04 & 0.967 & .50 & .40 & .41 & .06 \\
 & 35\% & 0.998 & .37 & .43 & .43 & .01 & 0.998 & .36 & .57 & .57 & .02 \\
\addlinespace
OLMo-2 & default & 0.500 & .65 & .01 & .11 & .18 & 0.500 & .70 & .16 & .20 & .21 \\
 & key-matched & 0.686 & .51 & .21 & .26 & .08 & 0.578 & .63 & .24 & .27 & .15 \\
 & 50\% & 0.699 & .51 & .23 & .27 & .07 & 0.706 & .50 & .40 & .41 & .07 \\
 & 35\% & 0.840 & .35 & .47 & .48 & .02 & 0.813 & .35 & .58 & .59 & .03 \\
\bottomrule
\end{tabular}}
\caption{The four judges at each operating point on PopQA, measured on the untrained student's 32 sampled answers to the training questions. \emph{Threshold}: an answer is accepted if P(Yes) exceeds it. \emph{Accepts}: share of all answers accepted. \emph{Forced out}: the least share of right answers that any grader accepting so few answers must reject. \emph{Right rejected} and \emph{wrong paid}: the judge's actual error rates. Right means accepted by the key or the audit: 65\% of Qwen's and 83\% of Llama's answers.}
\label{tab:judges}
\end{table*}

\paragraph{Strictness comes from the threshold.} Table~\ref{tab:judges} shows each judge at each cut. At the default threshold the judges accept 62--87\% of answers and pay 18--55\% of the wrong ones. Reaching 35\% acceptance takes thresholds of 0.97 to 0.998 for the Qwen, Mistral and Gemma judges: at that cut a rejection is usually an answer the judge itself rates as probably correct. The prompt does not offer the same control. Told to answer Yes only if the answer ``is fully correct and you are certain of it'', the Qwen judge at its default threshold rejects 10.4\% and 12.6\% of key-right PopQA answers (Qwen and Llama students), against 8.9\% and 11.2\% with the standard prompt, and still accepts 68\% and 76\% of all answers.

\paragraph{The judge knows most of what it rejects.} We asked the Qwen judge the 1{,}500 TriviaQA questions closed-book. Of the right answers it rejects at the 35\% cut, it answers the question correctly itself in 81\% of cases for the Qwen student's answers and 67\% for the Llama student's. Of the right answers it accepts, it answers 96\% and 93\%. Rejections are more frequent where the judge does not know the answer, and still two thirds to four fifths of them fall on questions it answers correctly.

\paragraph{Forced share.} If a grader accepts a share $A$ of the answers and a share $\rho$ of the answers is right, it must reject at least $\max(0,\, 1 - A/\rho)$ of the right answers, whatever it knows. This is the cost of strictness itself. A judge's \emph{blind spot} is what it does beyond this: which right answers it chooses to reject.

\paragraph{Graders with placed errors.} \emph{Random errors}: the answer key, with each right answer rejected and each wrong answer paid independently at fixed rates. \emph{Hash-placed paid strings}: a hash of the question and the normalised answer pays a fixed 35\% of wrong question--answer pairs, the same pairs at every visit. \emph{Relation types}: on PopQA, right answers are rejected with probability 0.8 on eight relation types and 0.2 elsewhere (visible), or on a random half of the questions of every type (hidden). \emph{Long answers}: the key pays right answers, and a wrong answer of at least five words is paid with probability 0.8. \emph{Copy of a judge}: right answers are rejected on exactly the training questions where a judge at its 35\% cut rejects the untrained answer; wrong answers are paid at random at that judge's rate. The \emph{scrambled} copy rejects the same number of questions chosen by a hash. The \emph{type-wide} copy rejects whole relation types, starting from those the judge rejects most, until the set is as large.

\paragraph{Strict graders with no judge.} \emph{Random rejections}: the answer key with right answers rejected at random at rates from 0.2 to 0.7; for each judge we compare with the three such runs nearest in held-out abstention. The rejections are drawn anew for every sampled answer, so no question carries a fixed verdict, and 17 of these 91 runs end in refusing nearly everything. The judge-free grader with fixed verdicts is the scrambled copy of a judge's rejections (Appendix~\ref{app:capitals}). \emph{Own vote}: the key for a question is the student's most frequent untrained answer if at least $k$ of its 32 samples give it, and \idk{} is the only paid reply otherwise; $k=32$ is the unanimous vote.

\paragraph{Jury, Sonnet, reference.} The jury accepts an answer if at least three of the five judges accept it, with the judges' cuts set so that the jury accepts 35\% of the untrained answers. Its verdicts are computed once on the untrained answers and used as a fixed key. As a classifier of right and wrong answers the jury is better than its members (AUROC 0.96 and 0.90 for the two students, against 0.90--0.94 and 0.82--0.87). Cut to accept 35\% it rejects eleven and ten of the twelve former-state training answers, like the single judges (7--12). Cut only until it pays as many wrong answers as single judges do at their 50\% cut, it rejects 16\% and 30\% of right answers where they reject 26--29\% and 41--43\%, and two and four of the twelve where they reject four to eight; we did not train with that looser jury. A single judge frozen in the same way reproduces live training with that judge (82\% of known former-state capitals refused in both cases, Qwen student). Sonnet (\texttt{claude-sonnet-5-5}) grades without a reference on a five-level scale and pays levels 4 and 5. The reference-guided judge is the Qwen judge with the reference answer in its prompt.

\section{Paid wrong answers: supporting results}
\label{app:hacks}

\paragraph{Paid answers come from the model's own samples.} With the hash grader we know exactly which wrong strings are paid. On training questions, a wrong string that the untrained model gave in 2 to 7 of its 32 samples becomes the final answer in 26.0\% (Qwen) and 20.7\% (Llama) of cases when the run's grader pays it, and in 0.5\% when it does not. Scored against another run's table, the same strings show no such gap (10.0\% against 9.6\%, and 8.3\% against 7.1\%). On held-out questions paid and unpaid strings are equally likely (9.8\% against 9.9\%, and 9.4\% against 8.9\%).

\paragraph{Visits, not steps.} We vary the number of training questions so that each is seen 1, 2, 5 or 10 times in the same 120 steps (TriviaQA, lenient Qwen judge, 15 runs per student). The first revisit raises the judge-paid share of wrong training samples by 0.12 in both students ([0.11, 0.13]); by the fifth visit the rise is 0.24--0.28. At a fixed number of visits, 100 more steps add 0.03 (Qwen) and 0.00 (Llama). Visits explain 15--18\% of the variance left by step position; steps explain 0.5\% of the variance left by visits.

\begin{table}[H]
\small
\centering
\setlength{\tabcolsep}{4pt}
\resizebox{\columnwidth}{!}{%
\begin{tabular}{@{}lrccc@{}}
\toprule
Cut & Runs & Training & Held-out & Ratio \\
\midrule
Default & 48 & 0.37 & 0.04 & 0.11 [0.07, 0.15] \\
50\% cut & 59 & 0.36 & 0.01 & 0.04 [$-$0.04, 0.11] \\
35\% cut & 66 & 0.30 & 0.03 & 0.12 [0.01, 0.24] \\
\bottomrule
\end{tabular}}
\caption{Extra pay a judge gives its own model's changed wrong answers, beyond what the other three judges give them, on training and on held-out questions, and the ratio of the two with its 95\% interval. Both students pooled.}
\label{tab:content}
\end{table}

\paragraph{Transfer at each operating point.} Table~\ref{tab:content} splits the 173 judge-trained runs by cut. The effect on training questions is 0.30--0.37 at every cut; on held-out questions it is 0.01--0.04. Against the other judges the pooled transfer is 0.09 [0.04, 0.14], the figure in the main text; against key-trained models it is 0.10 [0.05, 0.16]. Without PopQA's near-duplicate held-out questions the held-out effect is 0.024 [0.008, 0.042], against 0.029 with them.

\paragraph{What does carry over.} No run gives the same string on more than 2.1\% of held-out questions. Table~\ref{tab:scale} separates what one judge pays from what all judges pay: measured against the other judges, a judge's paid answers transfer at 0.09; measured against graders with random errors, which removes nothing the judges share, the figure is 0.26 [0.19, 0.36]. Most of that shared part is explained by the form of the answer and the kind of question (half to two thirds). The same holds for where a model answers: on questions where only its own judge pays the wrong answer, a model answers more, at 0.23 [0.09, 0.40] of the training effect at the 35\% cut and 0.35 [0.17, 0.55] at the default threshold. The extreme case is a form: in one of three runs, a Granite student trained without the KL penalty answers 590 of 718 held-out questions with one invented sentence that its judge pays.

\paragraph{The model's knowledge is untouched.} Under the KL penalty, 94.5\% of a trained model's right answers on held-out questions are the answers it gave before training. Of the accuracy gained among answered questions, 95\% comes from choosing which questions to answer and 5\% from changed answers.

\section{Blind spots: supporting results}
\label{app:blind}

\begin{table}[t]
\small
\centering
\begin{tabular}{@{}p{0.74\columnwidth}r@{}}
\toprule
Error & Transfer \\
\midrule
\multicolumn{2}{@{}l}{\emph{No rule to see, by construction: the scale's zero}} \\
\quad hash-placed paid strings & 0.01 \\
\quad rejections on a random half per type & 0.04 \\
\quad scrambled copy of a judge's rejections & $-$0.05 \\
\multicolumn{2}{@{}l}{\emph{Stays home: one judge's strings}} \\
\quad the wrong answers one judge pays & 0.09 \\
\multicolumn{2}{@{}l}{\emph{Travels: a pattern the model can see}} \\
\quad a strict judge's own rejections & 0.17--0.27 \\
\quad answering where one judge pays & 0.23--0.35 \\
\quad wrong answers every judge pays & 0.26 \\
\quad copy of a judge's rejections & 0.26 \\
\quad long answers paid & 0.67 \\
\quad rejections on visible relation types & 0.77--0.87 \\
\bottomrule
\end{tabular}
\caption{One scale for grader errors: the effect on held-out questions divided by the effect on training questions, pooled over both students. Errors placed by a rule no model could see fix the zero of the scale; whatever follows something visible in the question or the answer travels.}
\label{tab:scale}
\end{table}

\paragraph{Intervals for Table~\ref{tab:scale}.} Hash-placed strings 0.01 [$-$0.04, 0.07]; random half of each type 0.04 [$-$0.20, 0.35]; scrambled copy $-$0.05 [$-$0.18, 0.19]; wrong answers one judge pays 0.09 [0.04, 0.14]; a strict judge's own rejections 0.27 [0.21, 0.33], with two other estimators giving 0.17 and 0.22; copy of a judge's rejections 0.26 [0.15, 0.53]; long answers 0.67 [0.57, 0.76]; visible relation types 0.77 [0.62, 0.95], and 0.87 with a second estimator.

\paragraph{False rejections belong to the question.} For each question we take the distinct answers among the untrained student's 32 samples and each judge's score for them, for six judge configurations (the five judges and the Qwen judge with a strict prompt), two students and three datasets. Among right answers, the intraclass correlation of the score within a question is 0.61 (range over the 36 combinations 0.47--0.75); among wrong answers, 0.36 (0.18--0.49). For verdicts at the 35\% cut the figures are 0.53 for false rejections and 0.22 for false accepts, and the order holds in 36 of 36 combinations; it also holds when both error rates are set to 30\% (0.44 against 0.26, 36 of 36) or to 10\% (0.33 against 0.24, 32 of 36). Tightening a judge makes its verdicts more about the question: the share of verdict variance that lies between questions is 0.72 at the default threshold, 0.83 at the 50\% cut and 0.87 at the 35\% cut, rising in all 36 combinations, and at the 35\% cut it exceeds by 0.12 what a grader with the same error rates but answer-level errors would show.

\begin{table*}[t]
\small
\centering
\setlength{\tabcolsep}{8pt}
\begin{tabular}{@{}lccc@{}}
\toprule
 & Default & 50\% cut & 35\% cut \\
\midrule
Both students, all datasets & $+$0.03 [$-$0.00, 0.05] & $+$0.15 [0.11, 0.19] & $+$0.23 [0.20, 0.27] \\
Qwen student & $+$0.00 [$-$0.04, 0.04] & $+$0.12 [0.06, 0.17] & $+$0.21 [0.15, 0.27] \\
Llama student & $+$0.05 [0.01, 0.08] & $+$0.18 [0.12, 0.24] & $+$0.26 [0.21, 0.31] \\
TriviaQA & $+$0.06 [0.03, 0.08] & -- & $+$0.23 [0.17, 0.28] \\
PopQA & $-$0.01 [$-$0.07, 0.03] & $+$0.24 [0.17, 0.30] & $+$0.39 [0.31, 0.46] \\
NQ-open & $+$0.03 [$-$0.02, 0.08] & $+$0.06 [0.02, 0.10] & $+$0.15 [0.09, 0.21] \\
PopQA, Qwen student & $-$0.10 [$-$0.22, $-$0.01] & $+$0.17 [0.08, 0.27] & $+$0.31 [0.21, 0.41] \\
PopQA, Llama student & $+$0.03 [$-$0.03, 0.08] & $+$0.30 [0.20, 0.41] & $+$0.47 [0.37, 0.57] \\
\midrule
Judge pairs with a positive contrast & 19 of 31 & 21 of 24 & 43 of 44 \\
\bottomrule
\end{tabular}
\caption{The judge-against-judge contrast on held-out questions at each operating point: the refusal gap between right answers only judge $K$ rejects and right answers only judge $J$ rejects, in $K$'s models minus the same gap in $J$'s models. One script computes all three columns; the 35\% column agrees with the main-text estimate ($+$0.24 [0.19, 0.30]) up to the order in which students and datasets are averaged. TriviaQA has no runs at the 50\% cut for three of the judges.}
\label{tab:contrast}
\end{table*}

\paragraph{The contrast at each operating point.} Table~\ref{tab:contrast} gives the contrast by student and dataset. It is largest on PopQA and holds in every student--dataset cell at the 35\% cut. At the key-matched cut (Llama, PopQA) it is $+$0.11 [0.04, 0.18], positive in 6 of 6 pairs, so on this cell the ladder reads 0.03, 0.11, 0.30, 0.47. Adjusting for the student's confidence in each answer and for judge-free predictors of whether it is right leaves 0.21 [0.18, 0.25] of the pooled 0.23. Removing near-duplicate held-out questions leaves 0.21 pooled and 0.28 on PopQA. On PopQA, 37\% and 51\% of the contrast survive removing the mean refusal of each relation type (0.12 of 0.32 for Qwen, 0.23 of 0.46 for Llama): a judge's rejections follow relation types and also something finer than types.

\paragraph{Copies and placed rejections.} On held-out questions that the Mistral judge would reject, models trained with the copy of its verdicts show a larger refusal gap (rejected minus accepted answers, relative to judge-free models) than models trained with the scrambled copy, by 0.13 [$-$0.01, 0.24] (Qwen) and 0.50 [0.28, 0.74] (Llama), 0.32 [0.18, 0.44] pooled (exact permutation $p$ = 0.014 and 0.029). The copy and the live judge do not differ (0.00 [$-$0.06, 0.05]). Rejections placed on eight visible relation types carry 0.87 [0.72, 1.09] of their training effect to held-out questions; placed on a random half of every type, they leave 0.03 [$-$0.03, 0.12] on held-out questions that resemble the rejected ones. The rate matters: rejecting 80\% of a relation type's right answers moves refusal on new questions of that type by 0.39 [0.34, 0.46]; rejecting 50\% moves it by 0.10.

\paragraph{One pass is enough.} On TriviaQA we train the strict Qwen judge's models on 3{,}840 questions seen once (three runs per student) and on 750 questions seen about five times (six runs). The extra refusal of known answers that the judge rejects, beyond key-trained models, is 0.21 against 0.19 for Qwen and 0.25 against 0.29 for Llama.

\paragraph{The mirror image for wrong answers.} On NQ-open, where judges at the 35\% cut still pay enough wrong answers to test it, a model answers more on held-out questions where only its own judge would pay the wrong answer: the contrast is $-$0.15 [$-$0.24, $-$0.06], negative in 18 of 20 judge pairs. What travels is again a pattern over questions, here which questions are worth answering, and not the paid strings.

\paragraph{The loss can be forecast before training.} A linear probe on the untrained student's hidden states, fitted to a judge's verdicts on the training questions, predicts which held-out questions the trained model will refuse with AUROC 0.93 over 122 judge-trained runs. A probe that never sees the judge reaches 0.90, and the judge's own probe adds 0.02--0.04 in all 35 judge--student--dataset cells.

\section{The capitals probe}
\label{app:capitals}

\paragraph{Construction.} After seeing that judges reject right answers about former states on PopQA, we wrote a list of 270 former states and 65 present-day states by hand, with every commonly given capital accepted, before running any model or judge on it. The 65 present-day states are those whose capital question does not already occur in our data. Of the 270 former states, 124 have one accepted capital. Nothing is trained on the probe: the kept adapters of PopQA runs answer it. For each student we use the questions it knows before training (Llama 187 and 65, Qwen 149 and 63).

\begin{table*}[t]
\small
\centering
\setlength{\tabcolsep}{5pt}
\resizebox{\textwidth}{!}{%
\begin{tabular}{@{}llccccc@{\hspace{16pt}}ccccc@{}}
\toprule
 & & \multicolumn{5}{c}{Qwen student} & \multicolumn{5}{c}{Llama student} \\
\cmidrule(lr){3-7}\cmidrule(l){8-12}
 & & rejected & absten- & \multicolumn{2}{c}{model refuses} & judge-free & rejected & absten- & \multicolumn{2}{c}{model refuses} & judge-free \\
Grader & Cut & of 12 & tion & former & present & former & of 12 & tion & former & present & former \\
\midrule
Qwen3.5 & default & 1 & .07 & .00 & .00 & .01 & 2 & .01 & .00 & .00 & .03 \\
 & key-matched & -- & -- & -- & -- & -- & 6 & .26 & .42 & .00 & .05 \\
 & 50\% & 6 & .42 & .15 & .00 & .00 & 8 & .48 & .88 & .00 & .33 \\
 & 35\% & 12 & .63 & .82 & .04 & .18 & 11 & .61 & .98 & .01 & .54 \\
\addlinespace
Mistral-Nemo & default & 1 & .05 & .00 & .00 & .01 & 0 & .01 & .01 & .00 & .03 \\
 & key-matched & -- & -- & -- & -- & -- & 5 & .24 & .38 & .00 & .04 \\
 & 50\% & 5 & .43 & .29 & .01 & .01 & 7 & .36 & .51 & .00 & .17 \\
 & 35\% & 10 & .71 & .64 & .05 & .30 & 10 & .64 & .94 & .06 & .59 \\
\addlinespace
Gemma-4 & default & 1 & .13 & .00 & .00 & .01 & 2 & .04 & .00 & .00 & .03 \\
 & key-matched & -- & -- & -- & -- & -- & 2 & .10 & .03 & .00 & .03 \\
 & 50\% & 4 & .48 & .05 & .01 & .00 & 5 & .40 & .25 & .00 & .17 \\
 & 35\% & 7 & .60 & .56 & .03 & .14 & 8 & .57 & .91 & .01 & .43 \\
\addlinespace
OLMo-2 & default & 2 & .22 & .00 & .00 & .01 & 2 & .10 & .12 & .00 & .03 \\
 & key-matched & -- & -- & -- & -- & -- & 5 & .16 & .07 & .00 & .02 \\
 & 50\% & 7 & .44 & .28 & .01 & .00 & 5 & .38 & .65 & .00 & .15 \\
 & 35\% & 8 & .60 & .53 & .05 & .14 & 9 & .58 & .89 & .05 & .46 \\
\addlinespace
\multicolumn{2}{@{}l}{Answer key}  & 0 & .38 & .00 & .00 & .00 & 0 & .10 & .02 & .00 & .02 \\
\multicolumn{2}{@{}l}{Own vote, 21 / 29 of 32} & 2 & .31 & .13 & .00 & .01 & 2 & .41 & .34 & .00 & .21 \\
\multicolumn{2}{@{}l}{Own vote, 32 of 32} & 4 & .55 & .24 & .00 & .09 & 5 & .50 & .59 & .00 & .34 \\
\multicolumn{2}{@{}l}{Own vote, 20 of 32} & 1 & .38 & .09 & .00 & .00 & 0 & .19 & .11 & .00 & .02 \\
\bottomrule
\end{tabular}}
\caption{Every judge at every cut (the 28 points of Figure~\ref{fig:dose}), the answer key and the student's own vote. \emph{Rejected of 12}: how many of the twelve known former-state training answers the grader rejects. \emph{Abstention}: on PopQA held-out questions. \emph{Model refuses}: share of known former-state and present-day probe capitals refused. \emph{Judge-free}: former-state refusal of the three models trained with random rejections that are nearest in abstention. Two to six runs per cell. For Qwen the key-matched cut coincides with the 50\% cut. The 20-of-32 vote is the grader of Section~\ref{sec:own}, with all its runs at that cut (three with Qwen, six with Llama).}
\label{tab:dose}
\end{table*}

\begin{table*}[t]
\small
\centering
\resizebox{\textwidth}{!}{%
\begin{tabular}{@{}lccccc@{}}
\toprule
Grader (judges at the 35\% cut) & Grader accepts F / P & Model refuses F & Model refuses P & Beyond random rejections & Beyond own vote \\
\midrule
Qwen3.5 judge & .13 / .84 \quad .13 / .86 & .82 \quad .98 & .04 \quad .01 & $+$.61 \quad $+$.50 & $+$.58 \quad $+$.51 \\
Mistral-Nemo judge & .21 / .92 \quad .14 / .91 & .64 \quad .94 & .05 \quad .06 & $+$.31 \quad $+$.36 & $+$.39 \quad $+$.42 \\
Gemma-4 judge & .45 / .86 \quad .37 / .88 & .56 \quad .91 & .03 \quad .01 & $+$.36 \quad $+$.48 & $+$.33 \quad $+$.44 \\
OLMo-2 judge & .28 / .65 \quad .16 / .54 & .53 \quad .89 & .05 \quad .05 & $+$.36 \quad $+$.41 & $+$.28 \quad $+$.37 \\
Majority of five judges & & .85 \quad .95 & .14 \quad .04 & $+$.46 \quad $+$.37 & $+$.51 \quad $+$.45 \\
Copy of Mistral's verdicts & & .77 \quad .96 & .13 \quad .05 & $+$.36 \quad $+$.29 & $+$.43 \quad $+$.44 \\
Copy of OLMo's verdicts & & .63 \quad .84 & .08 \quad .14 & $+$.29 \quad $+$.25 & $+$.35 \quad $+$.24 \\
Scrambled copy of Mistral's verdicts & & .32 \quad .97 & .04 \quad .69 & $+$.03 \quad $-$.19 & $+$.07 \quad $-$.19 \\
Own vote, unanimous (no judge) & & .24 \quad .59 & .00 \quad .00 & $+$.16 \quad $+$.26 & \\
\bottomrule
\end{tabular}}
\caption{Capitals of former states (F) and present-day states (P), known answers, never trained. Each cell gives the Qwen student, then the Llama student. The last two columns are the refusal gap F minus P of the grader's models minus that of models trained with the key plus random rejections at matched abstention, or with the student's own vote. For the four judges and the jury every interval against both references excludes zero. The Llama entry of the scrambled copy includes one run that abstains on more than 95\% of held-out questions; without it the two shares are .95 and .60 (Table~\ref{tab:planted}).}
\label{tab:capitals}
\end{table*}

\begin{table*}[t]
\small
\centering
\resizebox{\textwidth}{!}{%
\begin{tabular}{@{}llcccc@{}}
\toprule
Judge & Student & Default threshold & Key-matched cut & 50\% cut & 35\% cut \\
\midrule
\multicolumn{2}{@{}l}{\emph{Right answers forced out}, Llama} & 0--3\% (16--19\% Gemma, OLMo) & 23--24\% & 39--41\% & 56--58\% \\
\multicolumn{2}{@{}l}{\emph{Right answers forced out}, Qwen} & 0--5\% & as 50\% & 23--26\% & 43--47\% \\
\midrule
Qwen3.5 & Llama & $-$.03 & $+$.38 [.19, .57] & $+$.55 [.42, .70] & $+$.50 \\
Mistral-Nemo & Llama & $-$.02 & $+$.33 [.19, .49] & $+$.34 [.19, .55] & $+$.36 \\
OLMo-2 & Llama & $+$.10 [.06, .14] & $+$.05 [.02, .10] & $+$.50 [.36, .63] & $+$.41 \\
Gemma-4 & Llama & $-$.03 & $+$.00 [$-$.03, .04] & $+$.08 [$-$.07, .22] & $+$.48 \\
Qwen3.5 & Qwen & $-$.01 & as 50\% & $+$.09 [$-$.04, .20] & $+$.61 \\
Mistral-Nemo & Qwen & $-$.01 & as 50\% & $+$.27 [$-$.01, .59] & $+$.31 \\
OLMo-2 & Qwen & $-$.01 & as 50\% & $+$.27 [.13, .43] & $+$.36 \\
Gemma-4 & Qwen & $-$.01 & as 50\% & $+$.04 [$-$.01, .11] & $+$.36 \\
\bottomrule
\end{tabular}}
\caption{The former-state blind spot at each operating point: refusal gap (former minus present-day) of the judge's models beyond models trained with the key plus random rejections at matched abstention. Two to six runs per cell.}
\label{tab:ladder}
\end{table*}

\paragraph{Every judge at every cut.} Table~\ref{tab:dose} gives the 28 points behind Figure~\ref{fig:dose}. The answer key rejects none of the twelve and its models keep the category (0--2\% refused). The student's own vote rejects none to five of the twelve, more as its cut rises, because these are answers the student is less sure of, and its models lose 9--59\%: this is the part of the effect that belongs to the student. At matched abstention the judges remove more than the own vote (Table~\ref{tab:capitals}), and the judge-free column shows how little of the loss abstention alone explains at the lower cuts.

\paragraph{The key-matched cut.} Here the comparison can be made at low abstention. The Llama models of the Qwen and Mistral judges abstain on 26\% and 24\% of held-out questions and refuse 42\% and 38\% of the former-state capitals (single runs 10--68\% and 24--67\%; five runs each). The student's own vote with a 20-of-32 rule rejects none of the twelve training answers, abstains on 19\%, and its models refuse 11\% over six runs (2\%, 5\% and 32\% in the first three, 9--10\% in the other three). Against the first three runs the judges' excess is 0.29 [0.07, 0.53] and 0.25 [0.03, 0.46]; they exceed random rejections at matched abstention by 0.38 [0.20, 0.55] and 0.33 [0.21, 0.48]. At this cut the blind spot is already present on average and not yet stable from run to run. For the Qwen student the key-matched cut is the 50\% cut. There the models of the Mistral and OLMo judges refuse 29\% and 28\% of the former-state capitals in two runs each (2\% and 56\%; 17\% and 39\%), those of the Qwen judge 15\% and of Gemma 5\%, while the student's own vote removes 9\% at a cut of 20 (2--13\% in three runs, abstention 0.38) and 24\% when all 32 samples must agree. Two runs per judge do not separate the judges from the own vote, so at this cut the result rests on Llama.

\begin{table*}[t]
\small
\centering
\setlength{\tabcolsep}{4pt}
\resizebox{\textwidth}{!}{%
\begin{tabular}{@{}lccccccc@{\hspace{14pt}}ccccccc@{}}
\toprule
 & \multicolumn{7}{c}{Qwen student} & \multicolumn{7}{c}{Llama student} \\
\cmidrule(lr){2-8}\cmidrule(l){9-15}
 & rejected & of other & & absten- & \multicolumn{2}{c}{model refuses} & random & rejected & of other & & absten- & \multicolumn{2}{c}{model refuses} & random \\
Grader (no judge in the loop) & of 12 & capitals & runs & tion & former & present & former & of 12 & capitals & runs & tion & former & present & former \\
\midrule
Type-wide copy of Mistral & 0 & .00 & 4 & .69 & .01 & .00 & .22 & 0 & .00 & 4 & .72 & .00 & .00 & .70 \\
Type-wide copy of OLMo & 0 & .00 & 4 & .70 & .02 & .00 & .22 & 0 & .00 & 4 & .66 & .01 & .00 & .70 \\
Scrambled copy of Mistral & 7 & .50 & 4 & .68 & .32 & .04 & .27 & 8 & .66 & 3 & .76 & .95 & .60 & .70 \\
Copy of OLMo & 8 & .14 & 4 & .69 & .63 & .08 & .33 & 9 & .21 & 4 & .60 & .84 & .14 & .47 \\
Copy of Mistral & 10 & .11 & 4 & .72 & .77 & .13 & .32 & 10 & .13 & 4 & .68 & .96 & .05 & .70 \\
Copy of Mistral, 0 of the twelve rejected & 0 & .11 & 2 & .67 & .15 & .01 & .18 & 0 & .13 & 2 & .62 & .31 & .01 & .57 \\
Copy of Mistral, 4 of the twelve rejected & 4 & .11 & 2 & .69 & .41 & .02 & .27 & 4 & .13 & 2 & .67 & .85 & .05 & .62 \\
Copy of Mistral, 8 of the twelve rejected & 8 & .11 & 2 & .71 & .81 & .06 & .27 & 8 & .13 & 2 & .64 & .91 & .02 & .57 \\
Copy of Mistral, 12 of the twelve rejected & 12 & .11 & 2 & .64 & .76 & .02 & .20 & 12 & .13 & 2 & .62 & .97 & .01 & .54 \\
\bottomrule
\end{tabular}}
\caption{Designed graders on the capitals. Each rejects right answers on a fixed set of training questions of about the same size and pays wrong answers at random at a judge's rate. The last four rows are the copy of the Mistral judge's rejections with the verdict on the twelve set by hand. \emph{Rejected of 12}: how many of the twelve known former-state training answers are in the set. \emph{Of other capitals}: share of the other known capital training answers in the set. \emph{Random}: former-state refusal of models trained with random rejections at matched abstention.}
\label{tab:planted}
\end{table*}

\paragraph{The same number of rejections, placed differently.} Table~\ref{tab:planted} holds the amount of rejection fixed and moves it. The type-wide copies reject whole relation types and none of the capital questions: their models abstain on 66--72\% of held-out questions and refuse 0--2\% of the former-state capitals, far below models trained with random rejections at the same abstention (22\% and 70\%). The copies of the Mistral and OLMo judges reject eight to ten of the twelve former-state answers and 11--21\% of the other capital answers: their models refuse 63--96\% of the former-state capitals and 5--14\% of the present-day ones. The scrambled copy rejects seven or eight of the twelve and half or more of the other capital answers: it produces no more of a boundary than random rejections do, and with Llama it removes capitals of both kinds (95\% and 60\%). What decides a category is whether its own training answers are rejected while its neighbours are not.

\paragraph{Changing the twelve verdicts alone.} The last four rows of Table~\ref{tab:planted} hold everything fixed except the twelve known former-state training answers. Each grader is the copy of the Mistral judge's rejections, with none, four, eight or all twelve of those answers in the rejection set; the four sets differ on at most twelve of 750 training questions, the share of other capital answers rejected stays at 11\% (Qwen) and 13\% (Llama), and held-out abstention stays at 62--71\%. Two runs were trained per grader and student. With Llama the models refuse 31\% (runs 9\% and 54\%), 85\% (79\%, 91\%), 91\% (89\%, 93\%) and 97\% (95\%, 98\%) of the unseen former-state capitals. With Qwen they refuse 15\% (12\%, 17\%), 41\% (34\%, 48\%), 81\% (78\%, 83\%) and 76\% (67\%, 85\%): the loss rises to eight rejected answers and is flat from eight to twelve. In both students every run with all twelve rejected lies above every run with none, and the difference of the means is 0.61 (Qwen) and 0.66 (Llama). Sparing the twelve leaves Qwen at the level of random rejections (15\% against 18\%) and Llama below it (31\% against 57\%), though one of the two Llama runs still loses 54\%; the type-wide copies, which reject no capital answer at all, lose 0--2\%. Present-day capitals are refused 1--6\% in every row.

\paragraph{The twelve verdicts at the key-matched cut.} We repeated the test where the grader is lenient. The grader copies the Qwen judge's rejections at the key-matched cut onto the training questions (186 or 198 of the 646 training questions for which Llama has a right answer, depending on the twelve; wrong answers are paid at the judge's rate, 38\%), with the twelve known former-state training answers either all accepted or all rejected. Three Llama runs were trained per grader. With the twelve accepted the models refuse 5\% of the unseen former-state capitals (runs 0\%, 1\% and 13\%); with the twelve rejected they refuse 90\% (93\%, 100\% and 79\%), and every run of the second kind is above every run of the first. Present-day capitals are refused by 0\% and 2\%. Of the 11 former-state capitals that Llama knows among the regular held-out questions, the runs with the twelve rejected refuse 11, 11 and 8, the others 0, 0 and 1. The models abstain on 8\% and 6\% of held-out PopQA questions, less than with the own vote. The reward by the answer key is 74.1 with the twelve accepted (runs 72.0, 74.8 and 75.4) and 71.4 with them rejected (71.1, 71.9 and 71.2).

\paragraph{An earlier probe: the kind is wider than former states.} Before the probe above we used every PopQA capital question outside our 1{,}500: 37 about former states, 30 about abolished administrative units and 249 about present-day places, most of them little known (median PopQA popularity 1{,}345, against 63{,}633 for our training questions). The students know only 5 (Qwen) and 12 (Llama) of the former-state answers there, which is why we built the larger probe. On the present-day places the students know (56 and 163), models of the strict Qwen judge refuse 61\% (runs 41--82\%) and 80\% (75--85\%), against 17\% and 63\% for random rejections at matched abstention, 1\% and 11\% for the answer key and 27\% and 44\% for the student's own vote. The 65 present-day states of the main probe are countries. Capitals of little-known present-day places are lost together with those of former states: what the judges doubt, and the models give up, is the capital of an entity that is not a familiar present-day country.

\paragraph{Controls.} Table~\ref{tab:capitals} sets the judges, the jury and the copies against both judge-free graders, and Table~\ref{tab:ladder} follows each judge across cuts. Across all judges and both students, no blind spot appears where the cut forces out under 5\% of the right answers, and every clear one sits at a cut that forces out at least 22\%. The scrambled copy makes the Llama student refuse former-state and present-day capitals alike (95\% and 60\%; 97\% and 69\% with a fourth run that refuses nearly everything): the same number of rejections, without the judge's choice of questions, produces abstention but no boundary. The gap is not produced by former states having several accepted capitals: restricted to the 124 with one, the Qwen student's judges still exceed matched random rejections by 0.31--0.58.

\begin{table*}[t]
\small
\centering
\setlength{\tabcolsep}{4pt}
\resizebox{\textwidth}{!}{%
\begin{tabular}{@{}llcccccc@{\hspace{14pt}}cccccc@{}}
\toprule
 & & \multicolumn{6}{c}{Qwen student} & \multicolumn{6}{c}{Llama student} \\
\cmidrule(lr){3-8}\cmidrule(l){9-14}
 & & \multicolumn{3}{c}{former-state} & \multicolumn{3}{c}{present-day} & \multicolumn{3}{c}{former-state} & \multicolumn{3}{c}{present-day} \\
\cmidrule(lr){3-5}\cmidrule(lr){6-8}\cmidrule(lr){9-11}\cmidrule(l){12-14}
 & & judge & model & accepted, & judge & model & rejected, & judge & model & accepted, & judge & model & rejected, \\
Judge & Cut & rejects & refuses & refused & rejects & refuses & refused & rejects & refuses & refused & rejects & refuses & refused \\
\midrule
Qwen3.5 & key-matched & -- & -- & -- & -- & -- & -- & .30 & .42 & .38 (131) & .05 & .00 & .00 (3) \\
 & 50\% & .59 & .15 & .09 (61) & .11 & .00 & .00 (7) & .64 & .88 & .80 (68) & .09 & .00 & .00 (6) \\
 & 35\% & .87 & .82 & .58 (20) & .16 & .04 & .13 (10) & .87 & .98 & .95 (25) & .14 & .01 & .09 (9) \\
\addlinespace
Mistral-Nemo & key-matched & -- & -- & -- & -- & -- & -- & .41 & .38 & .31 (110) & .03 & .00 & .00 (2) \\
 & 50\% & .45 & .29 & .24 (82) & .03 & .01 & .00 (2) & .66 & .51 & .33 (63) & .05 & .00 & .00 (3) \\
 & 35\% & .79 & .64 & .31 (31) & .08 & .05 & .33 (5) & .86 & .94 & .73 (27) & .09 & .06 & .22 (6) \\
\addlinespace
Gemma-4 & key-matched & -- & -- & -- & -- & -- & -- & .18 & .03 & .02 (154) & .03 & .00 & .00 (2) \\
 & 50\% & .27 & .05 & .04 (109) & .10 & .01 & .00 (6) & .29 & .25 & .18 (132) & .06 & .00 & .00 (4) \\
 & 35\% & .55 & .56 & .47 (67) & .14 & .03 & .11 (9) & .63 & .91 & .84 (69) & .12 & .01 & .04 (8) \\
\addlinespace
OLMo-2 & key-matched & -- & -- & -- & -- & -- & -- & .44 & .07 & .01 (104) & .11 & .00 & .00 (7) \\
 & 50\% & .42 & .28 & .18 (86) & .16 & .01 & .05 (10) & .63 & .65 & .41 (69) & .25 & .00 & .00 (16) \\
 & 35\% & .72 & .53 & .35 (41) & .35 & .05 & .11 (22) & .84 & .89 & .62 (30) & .46 & .05 & .08 (30) \\
\bottomrule
\end{tabular}}
\caption{The model against its own judge on the unseen probe. \emph{Judge rejects}: share of the known answers that the judge itself would reject at the cut. \emph{Model refuses}: share its trained models refuse. \emph{Accepted, refused}: of the former-state answers the judge accepts, the share the model refuses (number of such answers in brackets). \emph{Rejected, refused}: of the present-day answers the judge rejects, the share the model refuses.}
\label{tab:caric}
\end{table*}

\paragraph{The model against its own judge.} Table~\ref{tab:caric} gives the numbers behind Figure~\ref{fig:round}. Where a judge rejects most of a kind, its models refuse nearly all of it, including the answers the judge accepts. Where a judge rejects a minority of a kind, its models refuse almost none of it, including the answers the judge rejects: at the key-matched cut the OLMo judge rejects 44\% of Llama's former-state answers and its models refuse 7\%. With Llama the model rounds the judge's rate for a kind towards zero or one in every cell at the 35\% cut. With Qwen the model's refusal of former-state capitals stays below the judge's rejection rate in seven of eight cells, and it still refuses 31--58\% of the answers its judge accepts.

\paragraph{The judges that export nothing.} With Sonnet as the judge, no known capital of either kind is refused in any of six runs, and the models abstain as much as key-trained models on PopQA (38\% against 38\% for Qwen, 7\% against 10\% for Llama). The reference-guided Qwen judge at its default threshold gives a gap of $-$0.00 [$-$0.01, 0.01] (Qwen) and $-$0.02 [$-$0.06, $-$0.00] (Llama) beyond key-trained models. Held to the 35\% cut, the same reference-guided judge exports its own rejections: in all eight PopQA comparisons with reference-free judges each model refuses where its own judge rejects ($+$0.40 to $+$0.71). On the capitals its gap is $+$0.23 [$-$0.01, 0.51] and $+$0.70 [0.49, 0.90] beyond key-trained models, and $-$0.00 [$-$0.22, 0.29] and $+$0.22 [0.09, 0.49] beyond random rejections at matched abstention: smaller than a reference-free judge's.

\section{Kinds of question on fresh PopQA questions}
\label{app:kinds}

\begin{table*}[t]
\small
\centering
\setlength{\tabcolsep}{4.2pt}
\resizebox{\textwidth}{!}{%
\begin{tabular}{@{}lccccccc@{\hspace{14pt}}ccccccc@{}}
\toprule
 & \multicolumn{7}{c}{Llama student} & \multicolumn{7}{c}{Qwen student} \\
\cmidrule(lr){2-8}\cmidrule(l){9-15}
 & & rejected in & \multicolumn{4}{c}{refused on fresh questions by models of} & & & rejected in & \multicolumn{4}{c}{refused on fresh questions by models of} & \\
Kind (PopQA relation) & $n$ & training & judges & random rej. & own vote & key & excess & $n$ & training & judges & random rej. & own vote & key & excess \\
\midrule
screenwriter & 11 & .88 & 1.00 & .75 & .77 & .07 & $+$0.25 & -- & -- & -- & -- & -- & -- & -- \\
father & 11 & .66 & .96 & .64 & .50 & .05 & $+$0.32 & -- & -- & -- & -- & -- & -- & -- \\
country (``In what country is X?'') & 85 & .62 & .83 & .42 & .25 & .03 & $+$0.41 & 70 & .54 & .52 & .27 & .23 & .04 & $+$0.26 \\
genre & 22 & .61 & .74 & .89 & .93 & .00 & $-$0.15 & 24 & .60 & .76 & .94 & .60 & .33 & $-$0.18 \\
composer & 18 & .59 & .79 & .66 & .25 & .04 & $+$0.12 & -- & -- & -- & -- & -- & -- & -- \\
director & 27 & .55 & .73 & .59 & .30 & .02 & $+$0.15 & -- & -- & -- & -- & -- & -- & -- \\
occupation & 17 & .50 & .56 & .95 & 1.00 & .09 & $-$0.39 & -- & -- & -- & -- & -- & -- & -- \\
sport & 36 & .31 & .55 & .60 & .39 & .01 & $-$0.04 & 32 & .25 & .52 & .47 & .41 & .12 & $+$0.05 \\
author & 41 & .24 & .55 & .58 & .41 & .02 & $-$0.04 & 15 & .07 & .29 & .21 & .20 & .03 & $+$0.08 \\
capital & 26 & .18 & .87 & .45 & .38 & .06 & $+$0.42 & -- & -- & -- & -- & -- & -- & -- \\
\bottomrule
\end{tabular}}
\caption{Kinds of question on 2{,}569 PopQA questions that no model had seen (35\% cut; four judges averaged). $n$: fresh questions of the kind that the untrained student answers correctly in its greedy answer and in at least 31 of 32 samples. \emph{Rejected in training}: share of such answers on the training questions that the judges reject, counted before training. \emph{Refused}: share of the $n$ answers refused by the judges' models, by models trained with random rejections at matched abstention, with the student's own unanimous vote, and with the answer key. \emph{Excess}: judges minus random rejections. Kinds with at least 5 such training answers and 10 such fresh answers.}
\label{tab:kinds}
\end{table*}

\paragraph{The test.} The fresh questions are the 2{,}569 questions of a second PopQA sample that are not among our 1{,}500. They are spread over all six popularity bins of PopQA (median popularity 683, against 63{,}633 for our 1{,}500), so the test changes the popularity of the entities as well as the questions. No model was trained for this test: the kept adapters of 167 runs answered the questions, and the untrained students answered them with 32 samples. Right here means right by the answer key, since these answers were not audited. The count by kind was taken on the training questions before any output on the fresh questions existed. By the count alone the country kind is third of ten for Llama (62\% of sure, right training answers rejected; 54\% for Qwen). We singled it out because on its five held-out examples the judges' Llama models refused 53\% and judge-free graders none; the fresh questions are the test of that choice. The count itself needs no labels. Taken over the answers the student gives in at least 31 of its 32 samples, right or not (92\% of them are right for Llama, 89\% for Qwen), it ranks the relation types as the labelled count does (rank correlation 0.92 and 0.98), and the judges reject 75\% of such answers about former states, against 46\% and 39\% of all such answers.

\paragraph{What it shows.} Table~\ref{tab:kinds} gives every kind. On the country questions the judges' models refuse 83\% (Llama) and 52\% (Qwen) of the answers the student was sure of, 0.41 [0.27, 0.57] and 0.26 [0.16, 0.42] beyond random rejections (intervals over questions and runs; over questions alone [0.34, 0.49] and [0.19, 0.33]), for each judge separately ($+$0.31 to $+$0.58 with Llama, $+$0.15 to $+$0.40 with Qwen). Of the 84 country questions Llama answers correctly in all 32 samples, 56 are refused by all six models of the strict Qwen judge and by none of the five key-trained models. At the 50\% cut the excess is 0.13 [0.05, 0.20] and 0.12 [$-$0.05, 0.30], and at Llama's key-matched cut 0.01 [$-$0.07, 0.06]. The fresh capital questions, which are mostly about little-known places and former entities, are lost as the probe predicts (87\% against 45\% with Llama). The count taken in training ranks the kinds by what the judges' models refuse with Llama (rank correlation 0.65 at the 35\% cut and 0.50 at the 50\% cut over ten kinds); at the key-matched cut refusals are too few to rank (0.14), and with Qwen only four kinds have enough questions. The count does not say how much of a loss is the judge's own (rank correlation with the excess over random rejections 0.20 for Llama and $-$0.40 for Qwen), and a kind can sit inside a relation type: the relation \emph{capital} has the lowest count (18\%), because most training capital questions are about familiar countries, and the largest excess on the fresh questions. Elsewhere the judges remove less than strictness alone: on genre and occupation questions the judges' models refuse less than models trained with random rejections, which is the reallocation of Table~\ref{tab:realloc} seen on fresh questions.

\section{Shared doubts and aggregate numbers}
\label{app:shared}

\begin{table*}[t]
\small
\centering
\setlength{\tabcolsep}{3.6pt}
\resizebox{\textwidth}{!}{%
\begin{tabular}{@{}llccccc@{\hspace{12pt}}ccccc@{\hspace{12pt}}ccccc@{}}
\toprule
 & & \multicolumn{5}{c}{PopQA} & \multicolumn{5}{c}{TriviaQA} & \multicolumn{5}{c}{NQ-open} \\
\cmidrule(lr){3-7}\cmidrule(lr){8-12}\cmidrule(l){13-17}
 & & & \multicolumn{2}{c}{shared, of its} & \multicolumn{2}{c}{AUC from} & & \multicolumn{2}{c}{shared, of its} & \multicolumn{2}{c}{AUC from} & & \multicolumn{2}{c}{shared, of its} & \multicolumn{2}{c}{AUC from} \\
Student & Judge & rej. & rej. & acc. & judge & conf. & rej. & rej. & acc. & judge & conf. & rej. & rej. & acc. & judge & conf. \\
\midrule
Qwen & Qwen3.5 & .47 & .83 & .17 & .86 & .67 & .45 & .71 & .26 & .75 & .61 & .36 & .71 & .18 & .80 & .59 \\
 & Mistral-Nemo & .48 & .80 & .18 & .85 & .68 & .47 & .67 & .24 & .75 & .58 & .39 & .66 & .16 & .78 & .53 \\
 & Gemma-4 & .44 & .79 & .25 & .83 & .61 & .46 & .65 & .28 & .73 & .57 & .38 & .65 & .21 & .76 & .57 \\
 & OLMo-2 & .48 & .76 & .21 & .83 & .67 & .46 & .65 & .27 & .73 & .60 & .39 & .61 & .21 & .75 & .57 \\
\addlinespace
Llama & Qwen3.5 & .56 & .88 & .29 & .84 & .72 & .54 & .78 & .34 & .75 & .64 & .46 & .76 & .29 & .76 & .60 \\
 & Mistral-Nemo & .59 & .86 & .23 & .85 & .71 & .55 & .72 & .36 & .73 & .66 & .50 & .71 & .27 & .75 & .68 \\
 & Gemma-4 & .58 & .82 & .26 & .83 & .69 & .57 & .73 & .35 & .74 & .58 & .48 & .71 & .26 & .76 & .55 \\
 & OLMo-2 & .61 & .79 & .29 & .81 & .67 & .58 & .71 & .32 & .75 & .62 & .50 & .69 & .28 & .73 & .61 \\
\bottomrule
\end{tabular}}
\caption{Each judge at the 35\% cut on the right answers of the untrained student (training questions). \emph{Rej.}: share of right answers the judge rejects. \emph{Shared}: share of its rejections, and of its acceptances, that at least two of the other three judges also reject. \emph{AUC}: how well another judge's score (mean over the other three) and the student's confidence (agreement among its 32 samples) predict the judge's verdict.}
\label{tab:sharedfull}
\end{table*}

\paragraph{Every judge, every dataset.} Table~\ref{tab:sharedfull} gives every judge. The pattern holds for each judge with each student on each dataset: between 61\% and 88\% of a judge's rejected right answers are rejected by most of the other judges, against 16--36\% of its accepted ones. The student's confidence does not account for this. On PopQA the judges reject 35--37\% (Qwen) and 39--48\% (Llama) of the right answers the student gives 32 times out of 32.

\paragraph{A single judge's own part.} Inside the capitals probe, where one judge rejects an answer and another accepts it, the first judge's models refuse it more often by 0.11 [0.01, 0.22] (Qwen) and 0.17 [0.04, 0.30] (Llama) at the 35\% cut, and by 0.14 [0.03, 0.24] at the key-matched cut (Llama). The same questions are compared under two judges, so the student's own uncertainty cancels.

\begin{table}[H]
\small
\centering
\setlength{\tabcolsep}{3.5pt}
\resizebox{\columnwidth}{!}{%
\begin{tabular}{@{}lcccc@{}}
\toprule
Types by the judge's & Judge & \multicolumn{2}{c}{Refused by models of} & \\
acceptance (fifths) & accepts & the judge & judge-free & Excess \\
\midrule
\multicolumn{5}{@{}l}{\emph{35\% cut}} \\
\quad lowest & .10 & .93 & .87 & $+$.06 \\
\quad second & .30 & .72 & .65 & $+$.07 \\
\quad third & .43 & .54 & .61 & $-$.07 \\
\quad fourth & .59 & .37 & .60 & $-$.23 \\
\quad highest & .80 & .10 & .08 & $+$.02 \\
\multicolumn{5}{@{}l}{\emph{50\% cut}} \\
\quad lowest & .28 & .66 & .54 & $+$.12 \\
\quad second & .56 & .36 & .26 & $+$.09 \\
\quad third & .69 & .22 & .28 & $-$.06 \\
\quad fourth & .83 & .05 & .11 & $-$.05 \\
\quad highest & .89 & .04 & .02 & $+$.01 \\
\bottomrule
\end{tabular}}
\caption{A strict judge moves refusals between relation types. PopQA relation types sorted into fifths by how often the judge accepts the student's right answers of that type; share of known held-out answers refused by the judge's models and by models trained with random rejections at matched abstention. Four judges and both students pooled.}
\label{tab:realloc}
\end{table}

\paragraph{Refusals are moved between kinds.} Table~\ref{tab:realloc} sorts PopQA's relation types by the judge's acceptance. The judge's models refuse more than judge-free models on the types the judge accepts least ($+$0.07 [0.02, 0.11] on the lowest third at the 35\% cut, $+$0.11 [0.04, 0.18] at the 50\% cut) and less in the middle. Weighted by questions the excess is $-$0.01 and $+$0.02. A judge at these cuts adds no refusals; it decides where they go. The same holds over all held-out questions of the three datasets: at the 35\% cut the judges' models refuse 25.5 known answers per 100 questions, and models trained with random rejections, read at the same abstention off a line fitted to those runs in each student--dataset pair, refuse 26.6 (28.9 against 28.6 without Qwen on NQ-open, where the line rests on one run).

\begin{figure}[t]
\centering
\includegraphics[width=\columnwidth]{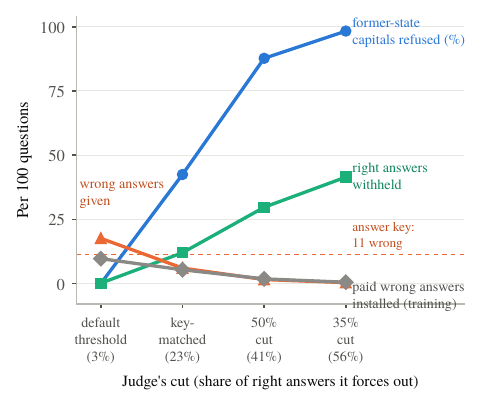}
\caption{Tightening one judge (Qwen judge, Llama student, PopQA). Paid wrong answers installed on training questions and wrong answers on new questions (audited truth) fall; right answers withheld and the former-state blind spot rise. No cut gives both.}
\label{fig:ladder}
\end{figure}

\begin{table*}[t]
\small
\centering
\setlength{\tabcolsep}{6pt}
\resizebox{\textwidth}{!}{%
\begin{tabular}{@{}llccc@{\hspace{18pt}}ccc@{}}
\toprule
 & & \multicolumn{3}{c}{Qwen student} & \multicolumn{3}{c}{Llama student} \\
\cmidrule(lr){3-5}\cmidrule(l){6-8}
Judge & Cut & paid wrong installed & right withheld & inherited & paid wrong installed & right withheld & inherited \\
\midrule
Qwen3.5 & default & 19.4 & 0.1 & .01 & 9.7 & 0.1 & .00 \\
 & key-matched & -- & -- & -- & 5.4 & 12.0 & .10 \\
 & 50\% & 2.7 & 9.6 & .12 & 1.8 & 29.7 & .21 \\
 & 35\% & 0.8 & 23.3 & .24 & 0.5 & 41.4 & .27 \\
\addlinespace
Mistral-Nemo & default & 19.3 & 0.0 & .00 & 7.4 & 0.1 & $-$.01 \\
 & key-matched & -- & -- & -- & 3.9 & 10.3 & .08 \\
 & 50\% & 2.4 & 9.1 & .05 & 1.3 & 20.1 & .06 \\
 & 35\% & 0.4 & 29.7 & .25 & 0.5 & 44.2 & .14 \\
\addlinespace
Gemma-4 & default & 17.7 & 0.9 & .00 & 10.5 & 0.7 & $-$.03 \\
 & key-matched & -- & -- & -- & 10.8 & 2.3 & .03 \\
 & 50\% & 2.7 & 12.0 & .16 & 2.8 & 23.5 & .21 \\
 & 35\% & 1.2 & 19.9 & .14 & 1.3 & 37.2 & .39 \\
\addlinespace
OLMo-2 & default & 6.6 & 1.3 & $-$.01 & 6.8 & 2.4 & .03 \\
 & key-matched & -- & -- & -- & 5.3 & 5.2 & .07 \\
 & 50\% & 3.1 & 8.6 & .07 & 2.9 & 21.2 & .13 \\
 & 35\% & 0.6 & 20.4 & .02 & 0.8 & 38.6 & .17 \\
\bottomrule
\end{tabular}}
\caption{The two costs of each judge at each cut on PopQA. \emph{Paid wrong installed}: judge-paid wrong answers that replace the untrained answer, per 100 training questions. \emph{Right withheld}: refusals of answers the model knows, per 100 held-out questions. \emph{Inherited}: share of the judge's own rejections of known answers on held-out questions that its model refuses beyond a judge-free model at matched abstention.}
\label{tab:costs}
\end{table*}

\paragraph{Both costs for every judge.} Table~\ref{tab:costs} extends Figure~\ref{fig:ladder} to all four judges and both students. Every judge follows the same course: at the default threshold it installs 7--19 paid wrong answers per 100 training questions and withholds almost nothing; at the 35\% cut it installs about one and withholds 20--44 known answers per 100 held-out questions.

\begin{table*}[t]
\small
\centering
\setlength{\tabcolsep}{4.5pt}
\resizebox{\textwidth}{!}{%
\begin{tabular}{@{}llcccc@{\hspace{14pt}}cccc@{}}
\toprule
 & & \multicolumn{4}{c}{Qwen student} & \multicolumn{4}{c}{Llama student} \\
\cmidrule(lr){3-6}\cmidrule(l){7-10}
Grader & Cut & right & wrong & \idk{} & return & right & wrong & \idk{} & return \\
\midrule
\multicolumn{2}{@{}l}{Untrained model} & -- & -- & -- & 0.653 & -- & -- & -- & 0.807 \\
\multicolumn{2}{@{}l}{Answer key} & .52 & .10 & .38 & 0.714 & .79 & .11 & .10 & 0.837 \\
\addlinespace
Qwen3.5 & default & .62 & .31 & .07 & 0.655 & .81 & .18 & .01 & 0.817 \\
 & key-matched & -- & -- & -- & -- & .68 & .06 & .26 & 0.810 \\
 & 50\% & .49 & .09 & .42 & 0.704 & .50 & .02 & .48 & 0.743 \\
 & 35\% & .35 & .02 & .63 & 0.669 & .38 & .00 & .61 & 0.690 \\
\addlinespace
Mistral-Nemo & default & .62 & .32 & .05 & 0.649 & .82 & .17 & .01 & 0.824 \\
 & key-matched & -- & -- & -- & -- & .70 & .06 & .24 & 0.819 \\
 & 50\% & .51 & .06 & .43 & 0.722 & .60 & .04 & .36 & 0.778 \\
 & 35\% & .29 & .01 & .71 & 0.641 & .36 & .00 & .64 & 0.677 \\
\addlinespace
Gemma-4 & default & .60 & .27 & .13 & 0.663 & .80 & .16 & .04 & 0.824 \\
 & key-matched & -- & -- & -- & -- & .77 & .12 & .10 & 0.825 \\
 & 50\% & .47 & .05 & .48 & 0.712 & .57 & .03 & .40 & 0.772 \\
 & 35\% & .39 & .02 & .60 & 0.685 & .43 & .00 & .57 & 0.713 \\
\addlinespace
OLMo-2 & default & .61 & .17 & .22 & 0.719 & .79 & .12 & .10 & 0.835 \\
 & key-matched & -- & -- & -- & -- & .76 & .09 & .16 & 0.835 \\
 & 50\% & .51 & .06 & .44 & 0.725 & .59 & .03 & .38 & 0.780 \\
 & 35\% & .38 & .02 & .60 & 0.684 & .41 & .01 & .58 & 0.704 \\
\addlinespace
\multicolumn{2}{@{}l}{Majority of five judges, 35\% cut} & .32 & .01 & .66 & 0.655 & .35 & .00 & .65 & 0.676 \\
\multicolumn{2}{@{}l}{Judge shown the reference, default} & .52 & .07 & .41 & 0.726 & .80 & .09 & .11 & 0.859 \\
\multicolumn{2}{@{}l}{Sonnet as judge} & .54 & .08 & .38 & 0.728 & .81 & .12 & .07 & 0.842 \\
\bottomrule
\end{tabular}}
\caption{What each grader's models earn on held-out PopQA questions under the training reward itself (1 for a right answer by key or audit, 0 for a wrong one, 0.5 for \idk{}), with the shares of right, wrong and refused answers.}
\label{tab:returns}
\end{table*}

\paragraph{What each cut earns.} Table~\ref{tab:returns} scores every model with the reward it was trained for, using audited truth. Tightening a reference-free judge cuts wrong answers at every step (31\% to 2\% for the Qwen student with the Qwen judge, 18\% to 0.4\% for Llama) and pays for them with refusals. For the Qwen student the 50\% cut earns the most with every judge (0.70--0.73); for Llama no cut earns clearly more than the default threshold, and the key-matched cut earns within a point of it (0.81--0.84) with fewer wrong answers (6--12\% against 12--18\%). The 35\% cut lowers the return of both. Pooled over the three datasets the pattern is the same for the Qwen judge (Qwen 0.60, 0.65, 0.64 at the default, 50\% and 35\% cuts; Llama 0.74, 0.72, 0.68). The judge shown the reference and Sonnet earn the most of all graders on PopQA, more than the answer key, at the abstention of key-trained models.

\section{The student's own vote as a yardstick}
\label{app:vote}
\label{app:sc}

\paragraph{The comparison.} A trained student abstains on some share of the held-out questions. The yardstick is the \emph{untrained} student following a rule: abstain on the same share of questions, taking first the ones it would not answer and then the ones where the fewest of its 32 samples agree with its usual answer (ties are shared), and give the usual answer everywhere else. Both are scored by true reward. The difference is positive when training with a grader chose what to answer better than the student's own vote does. It is not defined for the few runs that abstain less than the untrained student.

\begin{table*}[t]
\small
\centering
\setlength{\tabcolsep}{3.8pt}
\begin{tabular}{@{}lrccccccc@{}}
\toprule
 & & \multicolumn{3}{c}{Qwen2.5-7B student} & \multicolumn{3}{c}{Llama-3.1-8B student} & \\
\cmidrule(lr){3-5}\cmidrule(lr){6-8}
Trained with & Runs & PopQA & TriviaQA & NQ-open & PopQA & TriviaQA & NQ-open & Mean \\
\midrule
Answer key & 32 & $-1.2$ & $-0.8$ & $-0.1$ & $-0.1$ & $-1.3$ & $-1.3$ & $-0.8$ \\
Sonnet as judge & 18 & $-0.1$ & $-0.4$ & $+1.8$ & $+1.2$ & $-0.4$ & $-0.1$ & $+0.3$ \\
Judge shown the reference & 7 & $-0.2$ & -- & $+2.3$ & $+1.6$ & -- & $+0.1$ & $+1.0$ \\
Student's own vote (all cuts) & 49 & $-1.2$ & $-1.0$ & $-1.1$ & $-1.0$ & $-1.1$ & $-1.9$ & $-1.2$ \\
Judges, default threshold & 15 & $+2.9$ & $-1.0$ & $+0.3$ & $+0.2$ & $-2.9$ & $+2.4$ & $+0.3$ \\
Judges, 50\% or key-matched cut & 69 & $-1.0$ & $-0.3$ & $+0.2$ & $-1.1$ & $-0.7$ & $-0.2$ & $-0.5$ \\
Judges, 35\% cut & 83 & $+0.6$ & $-0.1$ & $+1.4$ & $+0.0$ & $-0.3$ & $+0.9$ & $+0.4$ \\
Key with random rejections & 66 & $-0.6$ & $-0.6$ & -- & $-0.9$ & $-1.0$ & $-0.8$ & $-0.8$ \\
\bottomrule
\end{tabular}
\caption{True reward of trained students minus the reward of the untrained student abstaining equally often by its own 32-sample vote (points), by family of graders. ``Judges'' are the reference-free judges of the students' size. Means over runs within a pair; the last column is the mean of the pairs.}
\label{tab:vote}
\end{table*}

\paragraph{Results.} Over the 584 runs for which the comparison is defined, the mean difference is $-$0.5 points and 34\% of the runs are above zero (Table~\ref{tab:vote}). Graders that know the answers stay within about a point of the rule: the answer key is below it in every pair ($-$0.8, 95\% interval $-$1.4 to $-$0.2 with the pairs as the sample), Sonnet 0.3 points above it on average ($-$0.6 to $+$1.3) and the judge shown the reference 1.0 above ($-$0.9 to $+$2.8, four pairs). Students trained with their own vote are 1.2 points below the rule they were trained to follow (1.4 at the cut of 20 used in Section~\ref{sec:own}).

\begin{table*}[t]
\small
\centering
\setlength{\tabcolsep}{4pt}
\resizebox{\textwidth}{!}{%
\begin{tabular}{@{}lcccccccccccc@{}}
\toprule
 & \multicolumn{3}{c}{Confident errors} & \multicolumn{3}{c}{Unsure errors} & \multicolumn{3}{c}{Confident right answers} & \multicolumn{3}{c}{Unsure right answers} \\
\cmidrule(lr){2-4}\cmidrule(lr){5-7}\cmidrule(lr){8-10}\cmidrule(l){11-13}
Trained with (pairs; runs) & right & dropped & wrong & right & dropped & wrong & right & dropped & wrong & right & dropped & wrong \\
\midrule
Answer key (6; 32) & 5 & 45 (26) & 50 & 7 & 68 (81) & 25 & 77 & 22 (16) & 1 & 48 & 44 (63) & 8 \\
Sonnet as judge (6; 18) & 6 & 24 (0) & 70 & 8 & 49 (59) & 42 & 93 & 6 (0) & 1 & 68 & 20 (36) & 12 \\
Judge shown the reference (4; 7) & 6 & 44 (7) & 50 & 8 & 70 (84) & 22 & 86 & 13 (2) & 1 & 63 & 30 (68) & 7 \\
OLMo judge, default (6; 10) & 7 & 19 (0) & 74 & 13 & 44 (55) & 44 & 94 & 4 (0) & 1 & 70 & 17 (24) & 13 \\
Student's own vote (6; 21) & 3 & 32 (7) & 66 & 3 & 66 (82) & 31 & 87 & 12 (4) & 1 & 57 & 35 (60) & 8 \\
\bottomrule
\end{tabular}}
\caption{What becomes of each of the untrained student's answers on held-out questions: the share that the trained student answers right, drops (answers with \idk{}), or answers wrong (\%; mean over runs, then over student--dataset pairs). An answer is \emph{confident} if at least 20 of the untrained student's 32 samples agree with it. In brackets: the share the rule drops when it abstains equally often. Collapsed runs are excluded.}
\label{tab:transshares}
\end{table*}

\begin{table*}[t]
\small
\centering
\setlength{\tabcolsep}{5pt}
\resizebox{\textwidth}{!}{%
\begin{tabular}{@{}lccccccc@{}}
\toprule
 & \multicolumn{3}{c}{Against the rule (points)} & \multicolumn{4}{c}{Changed, grader minus own vote} \\
\cmidrule(lr){2-4}\cmidrule(l){5-8}
Trained with & changed & dropped & total & points & pairs above zero & key as truth & no near-duplicates \\
\midrule
Answer key & $+$0.12 & $-$1.01 & $-$0.80 [$-$1.39, $-$0.21] & $+$0.37 [$+$0.11, $+$0.64] & 6 of 6 & $+$0.50 & $+$0.18 \\
Sonnet as judge & $+$0.50 & $-$0.31 & $+$0.33 [$-$0.64, $+$1.30] & $+$0.76 [$+$0.50, $+$1.02] & 6 of 6 & $+$0.47 & $+$0.58 \\
Judge shown the reference & $+$0.85 & $-$0.07 & $+$0.96 [$-$0.93, $+$2.84] & $+$1.25 [$-$0.02, $+$2.52] & 4 of 4 & $+$0.90 & $+$0.81 \\
OLMo judge, default & $+$1.45 & $-$0.89 & $+$0.71 [$-$2.39, $+$3.81] & $+$1.70 [$-$1.00, $+$4.41] & 5 of 6 & $+$1.32 & $+$1.43 \\
Student's own vote & $-$0.26 & $-$1.18 & $-$1.42 [$-$1.98, $-$0.87] & -- & -- & -- & -- \\
\bottomrule
\end{tabular}}
\caption{Where a trained student's true reward differs from the vote rule applied to the untrained student. \emph{Changed}: wrong answers turned right minus right answers turned wrong. \emph{Dropped}: wrong answers dropped beyond the rule minus right answers dropped beyond the rule, half a point each. The total also contains 0.0--0.2 points from questions the untrained student left unanswered. Brackets: 95\% $t$ intervals with the student--dataset pairs as the sample; the intervals of the other columns are in the released output. Last two columns: the same difference with truth taken from the answer key alone, and without held-out questions that are near-duplicates of a training question.}
\label{tab:transdec}
\end{table*}

\paragraph{Question by question.} Table~\ref{tab:transshares} follows each held-out question from the untrained student's answer to the trained student's, and Table~\ref{tab:transdec} splits the difference to the rule into answers changed and answers dropped. Every trained student changes a few answers in both directions. Students of the own vote, which was told no answers, correct 3.1\% of the untrained student's wrong answers and turn 1.8\% of its right answers wrong. Students of the answer key, Sonnet and the judge shown the reference correct 6.2\%, 7.5\% and 7.4\% and turn 2.1\%, 2.5\% and 2.0\% wrong. Net of the own vote, what these three graders know is worth 0.4, 0.8 and 1.2 points of changed answers, positive in every pair. The difference stays positive when truth is the answer key alone, so that Sonnet is not grader and auditor at once, and without near-duplicates of training questions. On PopQA the corrected answers are spread over relation types (father, director, screenwriter, capital of), and 43--83\% of them are near-duplicates, as are 43\% of all held-out questions.

In what they drop, trained students differ from the rule in one way: they drop more of the answers the untrained student was confident of, wrong and right alike, and keep more of the unsure ones. Students of the own vote do the same, so this comes from training and not from what a grader knows. In points it does not beat the rule for any grader. The OLMo judge's row is the change of form described next: its students correct 11\% of the wrong answers.

\paragraph{Where a grader does better: form.} Of the 173 combinations of grader and pair with at least two runs, 15 are more than one point above the rule and six more than two. The six are five cells of the OLMo judge, up to 5.3 points on NQ-open with Llama, and one of the grader that pays long answers. In these cells the answers changed shape. With the OLMo judge at the default threshold and at the 35\% cut, Llama's answers on NQ-open grow from 15 to 29 and 32 characters, 36\% and 38\% of them differ from the untrained answer, and 34\% and 37\% of the answered questions that the untrained student had wrong are now right by the audit (12\% with the answer key). We read fourteen of these answers: most are plain corrections of a year or a place inside a full sentence, and a few are answers that the audit accepts generously. The grader that pays long answers shows the same pattern (29 characters, 26\%). What helped is a form of answer, which the student can apply to any question, and not something the judge knew.

\paragraph{A probe gives the same reward.} We fit a logistic regression to the untrained student's hidden state at the last prompt token (layers 16 and 24, 256 principal components each) on the 750 training questions, with the audited correctness of the student's usual answer as the label, and use it as a rule: answer where the predicted probability of being right is at least one half. On new questions this rule earns 71.6 and 69.6 points on PopQA and TriviaQA for Qwen and 83.4 and 81.3 for Llama. RL with the answer key earns 71.4, 69.3, 83.7 and 81.0.

\paragraph{How much room is left.} A grader that paid exactly the right answers, followed perfectly, would earn 79.2, 80.6 and 69.8 points for Qwen on PopQA, TriviaQA and NQ-open, and 89.9, 89.1 and 79.1 for Llama. The best cut of the student's own vote, applied as a rule, earns 73.5, 71.3 and 59.5, and 84.9, 83.4 and 65.5. The rest would take knowledge the student does not have.

\section{The student's own vote as the grader}
\label{app:own}

\paragraph{Construction.} For each training question we group the untrained student's 32 sampled answers (two answers belong together if they are equal or one contains the other as whole words) and take the largest group. If it has at least $k$ answers, its answers become the question's key; otherwise the key is empty and only \idk{} is paid. Training then follows Appendix~\ref{app:train}. At $k = 20$ the share of training questions with an empty key is 40\%, 27\% and 56\% for Qwen on PopQA, TriviaQA and NQ-open, and 23\%, 21\% and 52\% for Llama.

\begin{table*}[t]
\small
\centering
\setlength{\tabcolsep}{5pt}
\begin{tabular}{@{}lccccrc@{}}
\toprule
 & Own vote & Answer key & Difference & & Wrong & Calibration \\
Pair & (3 runs) & (runs) & [95\% interval] & Abstains & per 100 & point \\
\midrule
Qwen, PopQA & 71.2 & 71.4 (6) & $-0.2$ [$-1.7$, $+1.3$] & 38\% & 10 & 13 \\
Qwen, TriviaQA & 65.9 & 69.3 (6) & $-3.4$ [$-5.8$, $-1.3$] & 12\% & 28 & 29 \\
Qwen, NQ-open & 57.7 & 55.1 (2) & $+2.6$ [$-1.0$, $+6.4$] & 49\% & 18 & 23 \\
Llama, PopQA & 82.9 & 83.7 (6) & $-0.7$ [$-2.3$, $+0.8$] & 20\% & 7 & 8 \\
Llama, TriviaQA & 80.5 & 81.0 (6) & $-0.5$ [$-1.9$, $+0.7$] & 19\% & 10 & 16 \\
Llama, NQ-open & 60.9 & 59.8 (6) & $+1.1$ [$-0.4$, $+2.6$] & 62\% & 8 & 14 \\
\midrule
Pooled & & & $-0.2$ [$-1.1$, $+0.7$] & & & \\
\bottomrule
\end{tabular}
\caption{The student's own vote at one cut (20 of 32 agreeing samples) against the answer key: true reward on new questions, the share of new questions on which the own-vote students abstain, their wrong answers per 100 new questions, and the student's calibration point. Intervals resample questions and runs. For Qwen on NQ-open four of the six answer-key runs answer \idk{} to every question; counting them at their reward of 50, the difference there is $+$6.0 and the pooled difference $+$0.4.}
\label{tab:own}
\end{table*}

\paragraph{One cut for all pairs.} Earlier own-vote runs used a different cut in each pair. From them we chose $k = 20$, fixed it for all six pairs, and trained three runs per pair; for Llama on PopQA, where three runs at this cut already existed, we used three new seeds (the earlier three score 83.5). Table~\ref{tab:own} gives the result. In five pairs the own vote is within 0.7 points of the answer key or above it. For Qwen on TriviaQA it is 3.4 points below, and the interval excludes zero. With the six pairs as the sample, the mean difference is $-$0.2 points with a 95\% $t$ interval from $-$2.3 to $+$1.9.

\paragraph{The category and fresh questions.} The own-vote students refuse 9\% of the former-state capitals they knew with Qwen (2\%, 13\% and 12\% in the three runs) and 11\% with Llama over its six runs at this cut (2\%, 5\% and 32\% in the earlier three; 10\%, 10\% and 9\% in the new ones), and none of the present-day capitals. A stricter vote loses more: with every sample required to agree, 24\% and 59\%. On the 2{,}569 fresh PopQA questions, scored by the answer key alone, the own-vote students earn 49.5 (Qwen) and 50.0 (Llama) points, key-trained students 50.0 and 48.6, students of the Qwen judge at its default threshold 34.3 and 31.1, and the untrained students 40.1 and 32.6.

\paragraph{Where the cut sits.} With a reward of 1, 0 and 0.5, an answer is worth giving when it is right more than half the time. We call a student's \emph{calibration point} the smallest number of agreeing samples at which its usual answer is right at least half the time, read from an isotonic fit on the training questions with audited labels (last column of Table~\ref{tab:own}). A cut below this point pays answers that are wrong more often than right. For Qwen on TriviaQA the answers just above the cut of 20, those with 20 to 25 agreeing samples, are right 47\% of the time; in the four pairs where the cut is above the calibration point they are right 65--83\% of the time. The student trained on this vote learns to answer nearly everything: it abstains on 12\% of new questions, where the rule at the same cut abstains on 29\% and earns 71.0 points, more than the key-trained students' 69.3.

Across all 20 settings of the cut that we trained (49 runs; Figure~\ref{fig:cut}), the 11 settings whose cut lies between two samples below the calibration point and twelve above it are between 0.9 points below the answer key and 3.0 above. The other nine are 3.0 points below on average, and the loss grows by about 0.8 points for every further sample of strictness. This range is read off the same runs and is a description of them. For Qwen on TriviaQA the two earlier runs with a cut of 27 are 0.4 points above the key.

\paragraph{Finding the calibration point takes labels.} Neither the judges nor the other student single out the overconfident pair: on the answers just above the cut, the Qwen judge accepts 77\% for Qwen on TriviaQA, where 47\% are right, and 59\% for Qwen on PopQA, where 65\% are right. With 100 labelled training questions the calibration point is estimated at 25 for Qwen on TriviaQA (5--95\% range over random draws 15--31) and with 5--95\% ranges that are 9 to 18 samples wide in the other pairs.

\section{What the students answer when they cannot refuse}
\label{app:forced}

\paragraph{Method.} Nothing is trained. A kept adapter answers each question greedily with the usual prompt. If the answer is a refusal (\idk{} or a variant such as ``Unknown''), the question is decoded again with the first token of that refusal blocked by a logit bias of $-100$, then with the first tokens of every refusal so far blocked, for up to five rounds. The first answer that is not a refusal is the \emph{forced answer}. An output is a refusal if it matches the pattern used throughout the paper or a wider list of refusal phrases, unless the answer key accepts it. We decoded 152 PopQA adapters on the capitals probe, and on the 1{,}500 questions of each dataset 118 adapters of five graders and the 8 kept adapters trained without the KL penalty, together with the untrained students. Outside collapsed runs, at most 0.1\% of the forced questions end without an answer.

\begin{table*}[t]
\small
\centering
\setlength{\tabcolsep}{4pt}
\begin{tabular}{@{}lcccc@{\hspace{12pt}}cccc@{}}
\toprule
 & \multicolumn{4}{c}{Qwen student} & \multicolumn{4}{c}{Llama student} \\
\cmidrule(lr){2-5}\cmidrule(l){6-9}
Trained with & runs & refusals & right & same & runs & refusals & right & same \\
\midrule
\addlinespace
\multicolumn{9}{@{}l}{\emph{Reference-free judges, 35\% cut}} \\
Qwen judge & 6 & 731 & 91 & 84 & 6 & 1{,}103 & 94 & 87 \\
Mistral judge & 3 & 288 & 89 & 84 & 3 & 526 & 94 & 87 \\
Gemma judge & 3 & 247 & 90 & 87 & 3 & 510 & 97 & 92 \\
OLMo judge & 3 & 236 & 89 & 85 & 3 & 499 & 97 & 91 \\
Jury of five judges & 4 & 511 & 82 & 75 & 4 & 710 & 93 & 87 \\
\addlinespace
\multicolumn{9}{@{}l}{\emph{Hand-set verdicts on the twelve (copy of Mistral)}} \\
\quad 12 rejected & 2 & 228 & 89 & 85 & 2 & 362 & 95 & 88 \\
\quad 8 rejected & 2 & 241 & 89 & 84 & 2 & 340 & 93 & 87 \\
\quad 4 rejected & 2 & 124 & 89 & 85 & 2 & 318 & 94 & 89 \\
\quad none rejected & 2 & 44 & 86 & 86 & 2 & 119 & 97 & 92 \\
\addlinespace
\multicolumn{9}{@{}l}{\emph{Reference-free judges, key-matched and 50\% cuts}} \\
Qwen judge, key-matched & -- & -- & -- & -- & 5 & 394 & 98 & 94 \\
Mistral judge, key-matched & -- & -- & -- & -- & 5 & 353 & 97 & 92 \\
Qwen judge, 50\% & 5 & 110 & 99 & 99 & 6 & 985 & 96 & 91 \\
Mistral judge, 50\% & 2 & 86 & 93 & 91 & 2 & 189 & 97 & 89 \\
Gemma judge, 50\% & 2 & 15 & 100 & 100 & 2 & 89 & 99 & 91 \\
OLMo judge, 50\% & 2 & 83 & 88 & 80 & 2 & 241 & 99 & 90 \\
\addlinespace
\multicolumn{9}{@{}l}{\emph{Other graders}} \\
Judge shown the reference, 35\% cut & 2 & 72 & 93 & 86 & 2 & 273 & 95 & 88 \\
Answer key, 40\% random rejections & 5 & 219 & 89 & 83 & 5 & 195 & 98 & 90 \\
Own vote, all 32 must agree & 2 & 70 & 89 & 86 & 2 & 221 & 96 & 90 \\
Own vote, 20 of 32 & 3 & 39 & 97 & 97 & 6 & 131 & 95 & 93 \\
\bottomrule
\end{tabular}
\caption{Former-state capitals of the probe that a student knew before training and refuses after it: what it answers once the refusal is blocked. \emph{Refusals}: number decoded, summed over runs. \emph{Right}: share of forced answers in the probe's answer list (\%). \emph{Same}: share equal to the answer of the untrained student (\%). Cells with fewer than five refusals are empty.}
\label{tab:forcedprobe}
\end{table*}

\begin{table*}[t]
\small
\centering
\setlength{\tabcolsep}{5pt}
\resizebox{\textwidth}{!}{%
\begin{tabular}{@{}lccccccc@{}}
\toprule
 & & & \multicolumn{3}{c}{Forced-choice accuracy (points)} & \multicolumn{2}{c}{On refused questions, right (\%)} \\
\cmidrule(lr){4-6}\cmidrule(l){7-8}
Trained with & Pairs; runs & Abstains (\%) & minus untrained & minus own vote & pairs above own vote & forced answer & untrained student \\
\midrule
Answer key & 6; 31 & 40 & $-$0.15 [$-$1.96, $+$1.66] & $+$0.72 [$-$0.96, $+$2.40] & 5 of 6 & 22.5 & 22.0 \\
Sonnet as judge & 6; 18 & 24 & $+$0.37 [$-$0.48, $+$1.22] & $+$1.24 [$-$0.01, $+$2.49] & 5 of 6 & 17.2 & 16.3 \\
Judge shown the reference & 4; 7 & 38 & $+$0.28 [$-$1.66, $+$2.23] & $+$1.44 [$-$0.02, $+$2.90] & 4 of 4 & 15.8 & 15.6 \\
Student's own vote & 6; 21 & 33 & $-$0.87 [$-$1.71, $-$0.03] & -- & -- & 20.7 & 22.5 \\
Qwen judge, 35\% cut & 6; 36 & 59 & $-$1.78 [$-$2.33, $-$1.23] & $-$0.91 [$-$2.05, $+$0.23] & 2 of 6 & 34.9 & 38.0 \\
\bottomrule
\end{tabular}}
\caption{Held-out questions with every refusal forced. \emph{Forced-choice accuracy}: share of questions whose final answer the answer key accepts, minus the same for the untrained student decoded in the same way (mean of student--dataset pairs; brackets: 95\% $t$ intervals over pairs; collapsed runs left out). Last two columns: on the questions a run refuses, how often its forced answer is right and how often the untrained student is right on the same questions.}
\label{tab:forcedheld}
\end{table*}

\paragraph{The capitals.} Pooled over the 15 runs of the four judges at the 35\% cut, the forced answer is right for 95.3\% of Llama's 2{,}638 refused former-state capitals (91--99\% across runs) and 90.0\% of Qwen's 1{,}502 (85--95\%); Table~\ref{tab:forcedprobe} gives every grader. Some forced answers that count as wrong are spellings outside the answer list (Samarqand, Kaegy\u{o}ng). The share is as high where few capitals are refused, at the key-matched cut and with the student's own vote, and with the hand-set graders.

\paragraph{Held-out questions.} Table~\ref{tab:forcedheld} scores all held-out questions. Students of the three graders that know the answers are within half a point of the untrained student. Students of the own vote are 0.9 points below it, and students of the Qwen judge at the 35\% cut 1.8 points below, in all six pairs. On the questions a student refuses, its forced answer is right about as often as the untrained student's. With key-or-audit truth the three differences are $-$1.8, $+$0.5 and $+$0.1 points; these understate the trained students, because 10--14\% of their final answers have no audit verdict and count as wrong, against 0.4--12\% for the untrained students.

\paragraph{Collapsed runs.} Five of the 118 runs answer \idk{} to at least 95\% of held-out questions: four answer-key runs and one run of the judge shown the reference, all with Qwen on NQ-open. Forced to answer, the four answer-key runs are right on 4.4\% of questions, against 26.0\% for the untrained student, and 45\% of their forced decodings end without an answer; the reference-judge run is right on 8.7\%. A collapsed student has lost answers as well as the will to give them. With these five runs counted, the answer key is 2.2 points below the untrained student and the judge shown the reference 1.8 below.

\paragraph{Without the KL penalty.} The eight kept adapters trained without the KL penalty (TriviaQA) abstain on under 2\% of questions. The two Llama runs with the lenient Qwen judge answer 38.6\% of held-out questions correctly, against 74.9\% for the untrained student; the two Qwen runs are 1.4 points above theirs. With an answer key that also pays 35\% of wrong answers, Llama is 2.8 points below and Qwen 2.0 above.

\section{Teaching abstention with the credit instead of the threshold}
\label{app:credit}

A judge can be left at its default threshold and the model taught to abstain by paying more for \idk{}. With group-normalised advantages the size of the credit cancels within a group, so these runs use unnormalised advantages (reward minus group mean) and a credit of 0.85 on PopQA.

\begin{table}[H]
\small
\centering
\setlength{\tabcolsep}{3.5pt}
\resizebox{\columnwidth}{!}{%
\begin{tabular}{@{}lcccc@{}}
\toprule
 & Abstains & Wrong & Right & Former \\
\midrule
\multicolumn{5}{@{}l}{\emph{Qwen student}} \\
\quad Answer key, credit 0.85 & .38 & .16 & .46 & .08 \\
\quad Qwen judge, credit 0.85 & .17 & .33 & .50 & .05 \\
\quad OLMo judge, credit 0.85 & .19 & .31 & .50 & .00 \\
\quad Qwen judge, 35\% cut & .63 & .04 & .33 & .82 \\
\multicolumn{5}{@{}l}{\emph{Llama student}} \\
\quad Answer key, credit 0.85 & .15 & .16 & .69 & .14 \\
\quad Qwen judge, credit 0.85 & .05 & .26 & .69 & .01 \\
\quad OLMo judge, credit 0.85 & .15 & .18 & .67 & .18 \\
\quad Qwen judge, 35\% cut & .61 & .03 & .36 & .98 \\
\bottomrule
\end{tabular}}
\caption{Abstention taught by a higher credit with judges at their default threshold, against the strict cut. Shares of PopQA held-out questions that are refused, answered wrongly and answered rightly, and the share of known former-state capitals refused. One or two runs per credit row. Right and wrong are by the answer key alone.}
\label{tab:credit}
\end{table}

The blind spot enters through the threshold, and the credit does not replace the threshold (Table~\ref{tab:credit}). With the credit raised, judge-trained models show no former-state gap beyond key-trained models at the same abstention ($-$0.06 and $-$0.11 for Qwen; $-$0.00 and $+$0.04 for Llama). But a lenient judge keeps paying wrong answers whatever \idk{} is worth: the Qwen student abstains half as often as with the key and gives twice as many wrong answers. Only the OLMo judge with the Llama student matches the key. A reference-free judge of this size must be made strict to teach abstention, and the threshold is the control that does it.

\section{Smaller students}
\label{app:small}

We repeat the main measurements with Phi-4-mini, SmolLM3 and a Granite student (446 runs; the numbers below are for TriviaQA). Paid wrong answers stay home: the extra pay a judge gives its own model's wrong answers is 0.25 on training questions and 0.00 on held-out questions for Phi-4-mini, 0.25 and 0.03 for SmolLM3, and 0.54 and 0.02 for the hash grader. Blind spots travel, more weakly than at 7--8B: the contrast between the Qwen and Gemma judges is $+$0.06 and $+$0.05. Knowledge is preserved less exactly than in the larger students: 86\% and 82\% of right held-out answers are the untrained answers. The remedy carries over: wrong strings installed per training question are 0.28 with reference-free judges, 0.05 with a reference-guided judge, and 0.01 with the answer key. The Granite student, trained with a judge that accepts free-form replies, is the one case where paid content reaches held-out questions (0.29 on training questions, 0.171 [0.035, 0.399] held-out), and what travels there is the form of the reply.

\section{Statistics}
\label{app:stats}

Intervals are 95\% percentile bootstraps with 1{,}000 to 2{,}000 resamples. Where runs and questions both vary we resample both: questions within student and dataset, and runs within grader. Contrasts between judges are also tested by permuting judge labels within cells, and comparisons of a few runs with a few runs by exact permutation of run labels. Analyses were developed on the Qwen student and repeated unchanged on Llama.

We designated ten primary tests. The jury, the self-consistency comparison, the fresh-question test and the graders with hand-set verdicts are not among them and rest on their own intervals. The ten are: paid wrong answers stay on training questions; they form on revisits; a lenient judge matches random errors on held-out questions; abstention follows the grader's leniency across students; the judge contrast; refusals follow what the student can predict of a judge; the copy beats the scrambled copy; visible placed rejections beat hidden ones; false rejections cluster by question more than false accepts; and the capitals gap. Pooled over the two students, all ten pass Holm correction (largest adjusted $p$ = 0.016). Within one student, two fall short of significance for each: for Llama, the equivalence test for paid wrong answers ($p$ = 0.13) and the copy against the scrambled copy (0.057); for Qwen, the lenient judge against random errors (0.12) and visible against hidden rejections (0.10).

\section{Reproducibility: every experiment and what it showed}
\label{app:repro}

\paragraph{From runs to numbers.} Each run saves its configuration, its training log, and the model's answer to every training and held-out question; the kept adapters answer the probes. Every number in the paper is computed from these files by a script, the figures and the per-run tables of this appendix are written by two scripts (\texttt{make\_figs.py}, \texttt{gen\_tables.py}) straight from the analysis outputs, and Tables~\ref{tab:capitals} and \ref{tab:ladder} are typed from the outputs named in the release. The release contains the training code, the run configurations, all saved answers, the judges' scores for every answer, the audit labels, the annotators' labels, the capitals probe with its answer list, and the analysis scripts with their outputs.

\paragraph{What was decided after looking.} We developed each analysis on the Qwen student and then ran it unchanged on Llama; both are always shown. The capitals probe was built after we saw in the training data that judges reject former-state capitals; its questions and answer list were written before any model or judge was run on them, its scoring was fixed on the Qwen judge, and it was then applied unchanged to the other judges, the jury, the copies, the reference-guided judge and Sonnet. The count of rejected training answers (Figure~\ref{fig:dose}), the comparison of a model with its own judge (Figure~\ref{fig:round}) and the shared rejections (Table~\ref{tab:sharedfull}) were computed after the runs existed, from quantities that are fixed before training. The four graders with hand-set verdicts on the twelve answers (Table~\ref{tab:planted}) were built, and their scoring script written, before those runs started. The cut of 20 for the student's own vote was chosen from earlier own-vote runs and fixed for all six pairs before the 18 runs of Section~\ref{sec:own}; the comparison of the cut with the calibration point (Figure~\ref{fig:cut}) was made after those runs, and its range is read off them. The comparison with the student's own vote at the same abstention (Figure~\ref{fig:vote}) and its question-by-question version (Table~\ref{tab:trans}) were computed after all runs existed. The forced decoding (Appendix~\ref{app:forced}) was run last; its scoring rules were fixed before any adapter was decoded. Table~\ref{tab:repro} lists every family of runs with its outcome, including the ones the main text does not use.

\begin{table*}[p]
\footnotesize
\centering
\begin{tabular}{@{}p{0.20\textwidth}rp{0.57\textwidth}p{0.11\textwidth}@{}}
\toprule
Experiment & Runs & What it showed & Where \\
\midrule
Answer key; key with random errors & 203 & The reference for every contrast. How often a model abstains follows how many answers its grader accepts; a curve fitted on Qwen predicts Llama ($R^2$ 0.85). & App.~\ref{app:train}, \ref{app:capitals}, \ref{app:shared} \\
Random grader at a judge's error rates & 18 & On new questions a lenient judge's models abstain like models of a random grader with the same error rates (difference $-$0.4 points [$-$1.5, 0.8]; Qwen $-$1.9, Llama $+$1.2). & this table \\
Reference-free judges, four cuts & 222 & Wrong accepts stay home; rejections travel; the capitals; shared rejections. & \S\ref{sec:home}--\ref{sec:travel} \\
Reference-guided judge; Sonnet & 34 & No blind spot and the highest reward; their students choose what to answer about as well as the student's own vote would, and correct about one more answer in a hundred than students of the own vote. Held to the 35\% cut the reference-guided judge exports its rejections. & \S\ref{sec:home}, \S\ref{sec:own}, App.~\ref{app:capitals}, \ref{app:shared}, \ref{app:sc} \\
Majority of five judges & 8 & Keeps the blind spot at the same strictness. & \S\ref{sec:travel} \\
Copies of a judge's verdicts & 91 & The copy reproduces the judge on held-out questions; the scrambled copy does not, and with Llama it removes capitals of both kinds (95\%, 60\%); the type-wide copy protects the capitals. Changing the copy's verdict on twelve training answers alone moves the loss of the category from 15--31\% to 76--97\% (16 runs), and at the key-matched cut from 5\% to 90\% (6 runs). A frozen copy of the Qwen judge's verdicts reproduces live training with that judge. & \S\ref{sec:travel}, App.~\ref{app:capitals} \\
Student's own vote & 50 & A grader with no labels and no judge. At one cut for all six pairs (18 runs) it is 0.2 points below the answer key pooled and within 0.7 points or above in five pairs; for Qwen on TriviaQA it is 3.4 points below and needs a stricter cut. It removes 9--11\% of the capitals, and fewer than any judge at matched abstention. & \S\ref{sec:own}, App.~\ref{app:own}, \ref{app:capitals} \\
Rejections placed on relation types or by popularity & 59 & Visible types transfer (0.87), a hidden half does not (0.03); 80\% rejection of a type moves new questions by 0.39, 50\% by 0.10. & App.~\ref{app:blind} \\
Hash-placed paid strings; long answers; any answer paid & 38 & Strings stay home (0.01); a paid form travels (0.67); paying any answer on a share of questions falls 5--8 points behind self-consistency. & \S\ref{sec:home}, App.~\ref{app:hacks}, \ref{app:vote} \\
Visits per question & 50 & Paid wrong answers form on revisits, not with steps; one pass is enough for a blind spot. & App.~\ref{app:hacks}, \ref{app:blind} \\
KL weight and run length & 29 & Without the KL penalty 22--94\% of answers on new questions differ from the untrained answer, against 8--17\% with it: the finding that knowledge is untouched belongs to this recipe. & this table \\
Credit with unnormalised advantages & 12 & A higher credit does not replace the threshold. & App.~\ref{app:credit} \\
Fresh PopQA questions, saved models & (167) & No training. A second category, named from the judges' rejections in training and five held-out examples, is lost on fresh questions (0.41 and 0.26 beyond random rejections); the count ranks the kinds for Llama at the 35\% and 50\% cuts (0.65, 0.50) and not at the key-matched cut (0.14). & \S\ref{sec:travel}, App.~\ref{app:kinds} \\
Open-ended requests with a refusal option & (0) & No training. A first step toward a second task: two untrained students answer 1{,}200 harmless and 500 harmful requests from ten categories, scored by two judges. The judges' rejections of harmless answers did not concentrate on categories (no category more than 0.10 above the median), so the task could not test the mechanism and no model was trained. & this table \\
Smaller students & 446 & Paid wrong answers stay home and rejections travel, more weakly; with one student trained without the KL penalty a paid form reaches held-out questions. & App.~\ref{app:small} \\
Forced decoding of kept adapters & (278) & No training. With \idk{} blocked, students of the strict judges give 95\% (Llama) and 90\% (Qwen) of the capitals they refuse correctly. Forced-choice accuracy of students trained with graders that know the answers is within half a point of the untrained student. Collapsed runs, and Llama trained without the KL penalty by a lenient judge, have lost answers. & \S\ref{sec:home}, \S\ref{sec:travel}, App.~\ref{app:forced} \\
\bottomrule
\end{tabular}
\caption{Every family of runs, the number of finished runs with the two main students (smaller students in the last row), and its outcome.}
\label{tab:repro}
\end{table*}

\paragraph{Claims that did not hold.} The release contains a ledger of the 95 claims we tested during analysis: 43 hold in both students, 33 hold pooled, in one student, or with a caveat that changes the wording, 10 fail, one is underpowered, and eight are magnitudes without a test. Seven further expectations were not met. The own vote at a single cut was to stay within 2.5 points of the answer key in every pair, and for Qwen on TriviaQA it is 3.4 points below (Appendix~\ref{app:own}). Judges on an open-ended refusal task were expected to reject harmless answers by category, which they did not (Table~\ref{tab:repro}). With none of the twelve training answers rejected, Llama was to refuse at most 25\% of the former-state capitals and refuses 31\%; the loss was to rise at every step, and for Qwen it is 81\% at eight rejected answers and 76\% at twelve (Appendix~\ref{app:capitals}). And the count taken in training was to rank the kinds at the cut where a student earns most (rank correlation of at least 0.5); at Llama's key-matched cut it is 0.14 (Appendix~\ref{app:kinds}). Forced-choice accuracy was to stay within two points of the untrained student for each grader with every run counted; with the five collapsed runs the answer key is 2.2 points below, and 0.15 below without them (Appendix~\ref{app:forced}). With the twelve verdicts set by hand at the key-matched cut, the reward by the answer key was to differ by at most 1.5 points between the two graders; it is 2.6 points lower with the twelve rejected (71.4 against 74.1, three runs each; Appendix~\ref{app:capitals}). The ledger entries that fail are these. A direct test that false rejections transfer more than false accepts because they are more predictable fails (difference in transfer 0.04 [$-$0.19, 0.27]); the data support the split between one judge's strings and what a question predicts, in both signs, and not a split by sign. Judges do not share false rejections more than false accepts once base rates are equalised. At the default threshold a judge's rejections leave no trace on held-out questions ($-$0.01 [$-$0.04, 0.02]; the estimate of Table~\ref{tab:contrast} for the same cut is $+$0.03 [$-$0.00, 0.05]). A hidden half of each relation type carries little, but equivalence to zero fails (0.03 [$-$0.03, 0.12]). An expected-reward rule with no fitted constants does not explain differences between judges ($R^2$ $-$0.68 at the 35\% cut). No grader beats 32-sample self-consistency in a way that replicates from Qwen to Llama. Without the KL penalty the answers do not stay the model's own (Table~\ref{tab:repro}). An earlier capitals probe had too few known former-state answers and showed that capitals of little-known present-day places are lost as well (Appendix~\ref{app:capitals}). The remaining failures follow. A curve that predicts abstention from the grader's leniency holds across the two students and for two of the three smaller students, and does not hold for runs that change the credit, the KL weight or the advantage estimator ($R^2$ $-$0.22), nor for the Granite student. A model of the abstention rate with no fitted constants predicted that forcing one \idk{} per group would lower abstention slightly; it rose by 7--9 points with Qwen. How predictable a judge's rejections are from the question did not rank how far they transfer within TriviaQA (rank correlation 0.31 [$-$0.18, 0.72]), and the Granite judge, the least predictable, shows the largest effect. On TriviaQA and PopQA strict judges pay too few wrong answers (1--6\%) to test whether one judge's false accepts leave a trace; NQ-open is where that test was possible (Appendix~\ref{app:blind}).

\paragraph{Per-student outcomes.} The ten tests of Appendix~\ref{app:stats} are reported for the two students together and separately, including the four that fall short of significance within one student. Cells have two to six runs, and where runs disagree the range is given (for example 10--68\% at the key-matched cut).

\end{document}